\documentclass[a4paper,fleqn]{cas-dc}

\usepackage{graphicx}
\usepackage[font=scriptsize, justification=centering]{caption}
\usepackage{subcaption}
\usepackage[linesnumbered, ruled,vlined]{algorithm2e}
\usepackage[switch]{lineno}
\usepackage{multirow}
\usepackage{enumitem}
\usepackage{listings}
\usepackage{xspace}
\usepackage{tcolorbox}
\usepackage{mdframed}
\usepackage[flushleft]{threeparttable}
\usepackage[numbers]{natbib}
\usepackage{color,soul}
\usepackage{xcolor}
\usepackage{booktabs, multirow}
\def\tsc#1{\csdef{#1}{\textsc{\lowercase{#1}}\xspace}}
\tsc{WGM}
\tsc{QE}
\tsc{EP}
\tsc{PMS}
\tsc{BEC}
\tsc{DE}
\newdefinition{definition}{Definition}
\definecolor{light-gray}{gray}{0.92} 
\newenvironment{gtheorem}%
{\begin{mdframed}[backgroundcolor=light-gray,
skipabove=5pt,
skipbelow=0pt,
nobreak=false
]\begin{mdtheorem}{name}{label}}%
{\end{mdtheorem}\end{mdframed}}
\definecolor{ao}{rgb}{0.0, 0.5, 0.0}
\newcommand{\tool}{\textsc{Canola}\xspace}
\newcommand{\semi}{\textsc{Docta}\xspace}
\newcommand{\semiv}{\textsc{Docta-V}\xspace}
\newcommand{\semir}{\textsc{Docta-R}\xspace}
\newcommand{\selc}{\textsc{SELC}\xspace}
\newcommand{\sidyp}{\textsc{SiDyP}\xspace}
\newcommand{\dividemix}{\textsc{DivideMix}\xspace}
\newcommand{\coteaching}{\textsc{Co-Teaching}\xspace}
\newcommand{\sceloss}{\textsc{SCE-Loss}\xspace}
\newcommand{\agnews}{\textit{AGNews}\xspace}
\newcommand{\fashion}{\textit{FashionMNIST}\xspace}
\newcommand{\organ}{\textit{OrganAMNIST}\xspace}
\newcommand{\resisc}{\textit{RESISC45}\xspace}
\newcommand{\yahoo}{\textit{Yahoo! Answer}\xspace}
\newcommand{\clothing}{\textit{Clothing1M}\xspace}
\newcommand{\etal}{\textit{et al.}\space}

\newcommand{\llm}{\textit{LLM}\space}
\newcommand{\superv}{\textit{Superv.}\space}
\newcommand{\semiSuperv}{\textit{Semi-Superv.}\space}
\newcommand{\weakSuperv}{\textit{Weak-Superv.}\space}

\begin{document}
\let\WriteBookmarks\relax
\def\floatpagepagefraction{1}
\def\textpagefraction{.001}

% Short title
\shorttitle{\tool}

% Short author
\shortauthors{Nguyen \textit{et~al.}}

\title [mode = title]{Noise-Aware Framework for Correcting Corrupted Labels}    

\author{Ha-Linh Nguyen}
[orcid=0009-0007-3748-4810]
%\ead{22024505@vnu.edu.vn}
\affiliation{organization={Faculty of Information Technology, VNU University of Engineering and Technology},
    city={Hanoi},
    country={Vietnam}}

\author{Hong-Anh Nguyen}
[orcid=0009-0005-9294-3763]
%\ead{23021466@vnu.edu.vn}

\author{Minh-Duc La}
[orcid=0009-0003-8419-9503]
%\ead{23020526@vnu.edu.vn}

\author{Phong Lam}
[orcid=0009-0002-5745-3361]
%\ead{22028164@vnu.edu.vn}

\author{Thu-Trang Nguyen}
[orcid=0000-0002-3596-2352]
\ead{trang.nguyen@vnu.edu.vn}
\cormark[1]

\author{Son Nguyen}
[orcid=0000-0002-8970-9870]
%\ead{sonnguyen@vnu.edu.vn}

\author{Hieu Dinh Vo}
[orcid=0000-0002-9407-1971]
%\ead{hieuvd@vnu.edu.vn}

% Corresponding author text
\cortext[cor1]{Corresponding author}

% Here goes the abstract

\begin{abstract}
High-quality labeled data is essential for training reliable ML/DL models. However, real-world datasets often contain a considerable proportion of corrupted labels, which can severely degrade model performance. 
To address this problem, we propose \tool, a novel framework for correcting corrupted labels through noise-aware learning and iterative label refinement. 
\tool explicitly estimates the underlying noise distribution of the dataset and incorporates this information into the training of a noise-aware Deep Neural Network. By incorporating noise characteristics during learning, \tool enables the model to down-weight unreliable supervision signals and focus on trustworthy patterns, thereby improving robustness and generalization. Label correction is performed via cautious, iterative soft label refinement, in which model predictions are blended with observed labels to prevent premature or erroneous updates. This progressive refinement allows the dataset to be repaired in a stable and controlled manner.
We evaluate \tool on six widely used datasets under realistic noisy labeling scenarios. Experimental results show that \tool consistently outperforms SOTA label correction methods, achieving relative improvements ranging from 19\% to 52\% in error reduction. Moreover, models trained on datasets corrected by \tool obtain substantial downstream performance gains. Even simple classifiers trained on \tool's corrected data can outperform complex model-centric approaches by margins of up to 67\%.

\end{abstract}


\maketitle

\section{Introduction}

%the problem of automated data labeling and noise labels
High-quality data is the foundational resource for training effective Machine Learning (ML) and Deep Learning (DL) models. In particular, supervised learning, the most widely adopted training paradigm~\cite{jordan2015machine}, heavily relies on large volumes of accurately annotated data to achieve reliable performance. 
However, real-world datasets, whether manually labeled by humans or automatically annotated using labeling tools~\cite{alchemist, desmond2021semi, zhu2024apt}, often contain a considerable proportion of corrupted labels~\cite{chenglearning, song2019selfie, clothing1M}.
Such mislabeled instances can significantly degrade model generalization capability, leading to unreliable or biased predictions.

%existing approach: model centric vs data centric and limitations of the existing approaches
To address the challenge of label noise, prior research has typically followed two main directions: \textit{model-centric} and \textit{data-centric}. 
Model-centric approaches~\cite{chenglearning, sceloss, dividemix, han2018co, khoshgoftaar2010comparing, li2019learning} focus on enhancing algorithmic robustness of ML/DL models by designing model architectures or learning algorithms that reduce overfitting to erroneous supervision. 
Meanwhile, data-centric approaches~\cite{simifeat, kim2024learning, cola, retrieval-based, noiserank, sidyp, selc, wang2024noisegpt} aim to improve the quality of the training data itself by detecting and/or correcting mislabeled instances. 
Li~\etal~\cite{li2021cleanml} empirically demonstrated that by directly targeting the root cause of label corruption, data-centric methods often yield superior downstream performance compared to model-centric techniques.

Following the data-centric perspective, a variety of methods~\cite{simifeat, kim2024learning, cola, retrieval-based, noiserank} have been proposed to detect corrupted labels. 
While these approaches can effectively identify which instances are potentially mislabeled,
they typically stop at detection. As a result, developers are still required to manually inspect and correct the mislabeled instances, which is labor-intensive and error-prone at scale.

Despite the importance of \textit{automatically repairing corrupted labels}, this problem remains relatively underexplored.
The limited existing label correction studies~\cite{simifeat, sidyp, selc} can be broadly categorized into two groups based on the source of their correction signals: (1) \textit{neighbor-based correction}, which relies on neighboring labels for correction, and (2) \textit{model-based correction}, which leverages the model's own predictions to refine noisy labels during training.

Neighbor-based correction approaches infer repaired labels by utilizing the labels of similar samples. For example, \semi~\cite{simifeat, docta} detects mislabeled samples by identifying label inconsistencies within their neighborhoods, and corrects them using majority voting or ranking strategies. Similarly, \sidyp~\cite{sidyp} generates candidate labels from neighboring instances and employs a diffusion model to refine these candidates. However, these neighbor-based correction methods rely on the assumption that similar instances are likely to share the same label.
While these approaches are effective when local neighborhoods are reliable, they struggle near decision boundaries, where visually or semantically similar instances may legitimately belong to different classes.

Model-based correction approaches~\cite{selc, zhang2025efficient} directly use the model's own predictions as correction signals.
Specifically, these methods progressively update training labels by blending the original annotations with predictions obtained during early training stages. This strategy is motivated by the empirical observation that Deep Neural Networks (DNNs) trained on noisy datasets tend to learn clean and easy patterns first, before gradually memorizing hard and noisy samples~\cite{arpit2017closer}. Consequently, predictions from early training iterations are assumed to better reflect the underlying clean label distribution and are used to correct mislabeled instances.
Despite their effectiveness, these approaches suffer from two major limitations.
First, because labels are typically updated at every training epoch, the correction process can be highly sensitive to fluctuations in model predictions.
Second, in realistic settings with high noise levels (where noisy samples dominate) or in the presence of ``hard-but-clean'' and ``easy-but-noisy'' instances, early-stage predictions are not necessarily reliable. 
Consequently, this triggers a confirmation bias loop, where the model propagates and amplifies its own incorrect predictions, causing erroneous label updates that compound the noise rather than reduce it.

%our approach
This paper introduces \tool, a novel framework for robust and stable corrupted label correction. Motivated by the strong generalization capability of DNNs~\cite{rolnick2017deep, zheng2020error}, \tool rethinks the timing and reliability of the correction signal.
% also leverages model predictions as informative signals to guide label refinement.
%
Different from prior approaches~\cite{selc, zhang2025efficient} that interleaves label updates with early-stage learning, making the correction process highly sensitive to unstable or erroneous predictions, \tool decouples the learning phase from the correction phase to ensure label updates are cautiously guided by a mature, converged model state.
%
% \tool focuses on learning a reliable model and performing cautious label refinement.

Specifically, to ensure the reliability of model predictions, \tool first trains a noise-aware DNN that is regularized by an estimated noise distribution of the dataset.
The core idea is that \textit{by incorporating noise characteristics into the learning process, the model can down-weight unreliable signals and focus on trustworthy patterns, thereby improving robustness and generalization}. 
Additionally, to mitigate premature or erroneous updates, \tool employs a loss-stabilization trigger by performing label refinement only after the training trajectory has stabilized, ensuring that model predictions are sufficiently reliable.
Label refinement is conducted through iterative soft relabeling, in which model predictions are blended with the observed labels.
These refined labels are then used in subsequent iterations, enabling the dataset to be progressively repaired in a stable and controlled manner.

%results
To evaluate the effectiveness of our proposed approach, we conduct extensive experiments on six widely used datasets covering both image and text classification tasks. Unlike prior work~\cite{simifeat, cola, selc, zhang2025efficient}, which primarily experiments under idealized synthetic noise that is often overly simplistic and may fail to reflect the complexity of real-world label noise, our evaluation is conducted under more realistic noisy conditions.
Specifically, we construct experimental data by applying practical labeling pipelines using a diverse set of automated data annotation techniques~\cite{alchemist, desmond2021semi, zhu2024apt, zhu2024coral, wang2021want}.  
Moreover, we also evaluate our approach on a real-world noisy dataset, \clothing~\cite{clothing1M}.

Our experimental results show that \tool consistently outperforms the baseline methods across all datasets and noise settings.
On average, it reduces the original dataset error rates by approximately 25\%, with substantially larger improvements under high-noise conditions. Compared with the state-of-the-art (SOTA) corrupted label correction techniques, \tool achieves relative improvements ranging from 19\% to 52\% across different settings. 
Moreover, by effectively improving the label quality of the training data, \tool leads to significant improvements in downstream model performance. In particular, under severe noise scenarios, models trained on data corrected by \tool outperform the model-centric approaches~\cite{sceloss, dividemix, han2018co, sidyp} by margins of 8\% to 67\%.
These results highlight that explicitly cleaning corrupted labels is more effective than relying solely on noise-robust training algorithms.

% Contributions
In brief, this paper makes the following contributions:

\begin{itemize}
    \item We construct a realistic benchmark consisting of five datasets designed to facilitate the evaluation of  corrupted label detection and correction methods under practical labeling scenarios. These datasets are constructed using diverse automated data annotation techniques.
    
    \item We introduce \tool, a novel corrupted label correction framework that decouples feature learning from label refinement. By combining a noise-aware DNN with a loss-stabilized trigger, \tool leverages the model's predictions to perform cautious, iterative label refinement.
    
    \item We conduct extensive experiments demonstrating that \tool consistently outperforms SOTA label correction methods. The datasets corrected by \tool also enable downstream classifiers to outperform model-centric approaches.
\end{itemize}

\section{Problem Formulation and Approach Overview}

\subsection{Problem Formulation}
We consider a standard classification task with $C$ classes.
Let $\mathcal{X}$ denote the feature space and $\mathcal{Y} = \{1, \dots, C\}$ be the label space. 
A clean dataset is defined as $D = \{(x_i, y_i)\}_{i=1}^{N}$, where each instance $(x_i, y_i)$ is independently and identically drawn from the joint distribution over $\mathcal{X} \times \mathcal{Y}$. Here, $x_i \in \mathcal{X}$ represents the feature vector and $y_i \in \mathcal{Y}$ denotes its corresponding \textit{clean} label.

In real-world scenarios, clean/true labels are not always available. Instead, we typically have access to a \textit{noisy labeled dataset} $\widetilde{D} = \{(x_i, \widetilde{y}_i)\}_{i=1}^{N}$, where each observed label $\widetilde{y}_i$ may differ from the true label $y_i$. An instance $(x_i, \widetilde{y}_i)$ is considered \textit{corrupted} if $\widetilde{y}_i \neq y_i$, and clean otherwise.
\textit{The goal of corrupted label correction is to identify instances with unreliable labels and replace them with more accurate estimates.} 
Formally, corrupted label correction approaches aim to transform the noisy dataset $\widetilde{D}$ into a corrected one $D^* = \{(x_i, y_i^*)\}_{i=1}^N$, where $y_i^*$ denotes an estimated clean label that ideally matches the true label $y_i$.
%

% To comprehensively evaluate the effectiveness of corrupted label correction methods, we adopt two complementary evaluation metrics: (i) \textit{Dataset Error Rate} and (ii) \textit{Downstream Model Performance}.
% \textit{Dataset Error Rate} measures the proportion of incorrect labels remaining in the corrected dataset $D^*$. This reflects the precision of the label repair process. A lower error rate indicates more accurate label recovery.
% \textit{Downstream Model Performance} assesses the practical utility
% of  $D^*$ for the downstream tasks. This is evaluated by training a classifier on $D^*$ and evaluating its performance, e.g., classification accuracy. Higher downstream performance implies that the corrected dataset $D^*$ better support effective model training.

\subsection{Approach Overview}
\begin{figure*}[t]
    \centering
    \includegraphics[width=\linewidth]{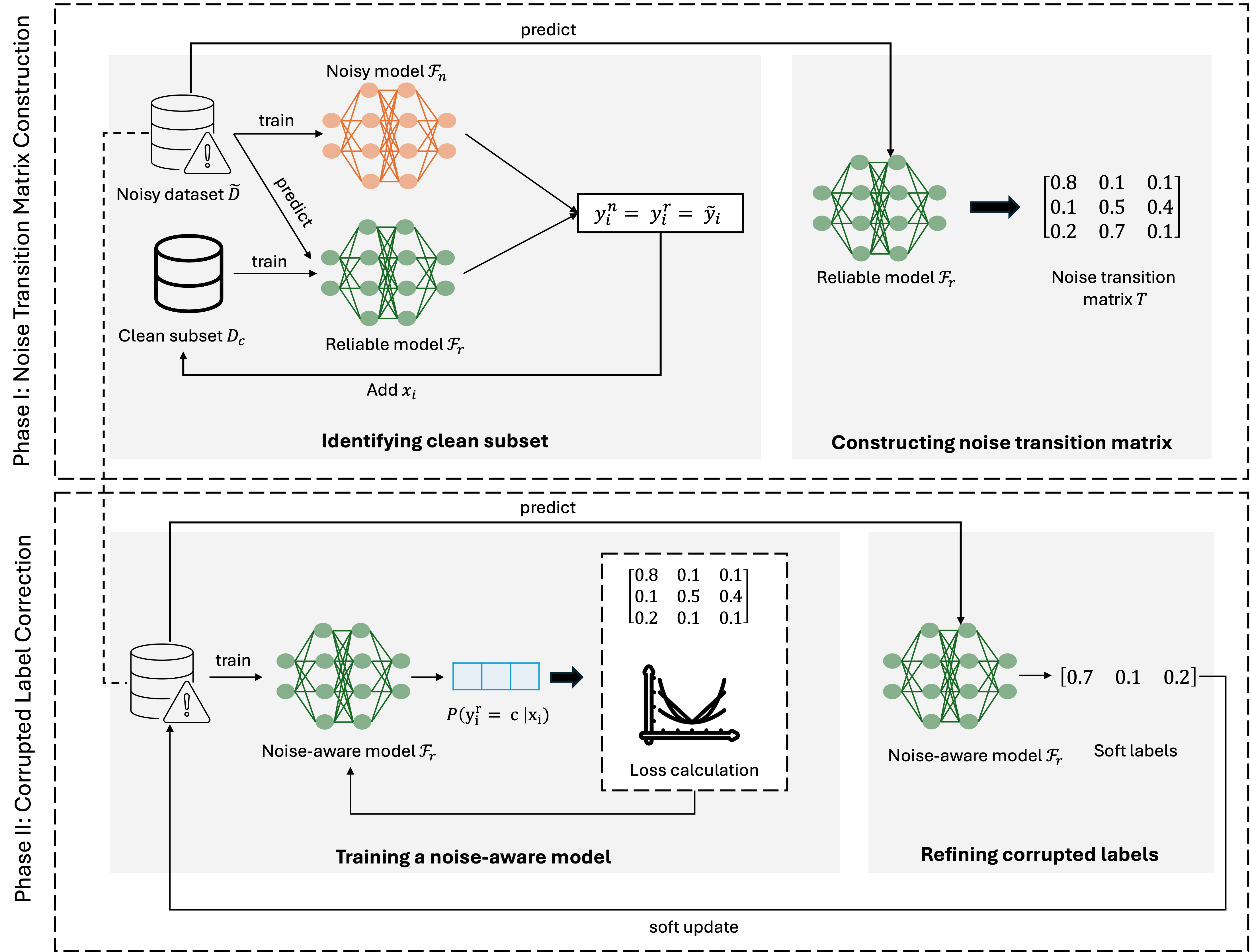}
    \caption{Overview of \tool: An iterative pipeline alternating between \textit{Noise Transition Matrix Construction (\textbf{Phase~1})} via asymmetric co-training and \textit{Corrupted Label Correction (\textbf{Phase~2})} using noise-aware learning and loss-triggered soft relabeling}
    \label{fig:pipeline}
\end{figure*}

Motivated by the intrinsic capability of DNNs to capture and generalize meaningful patterns even in the presence of label noise~\cite{rolnick2017deep, zheng2020error}, we leverage model predictions to guide the repair of corrupted labels. However, naive reliance on raw predictions often causes error propagation and unstable corrections.
To address these challenges, we propose \tool, a novel framework for corrupted label correction that decouples noise estimation from label refinement training a noise-aware model and performing cautious label refinement.
Figure~\ref{fig:pipeline} illustrates the overall workflow of \tool, which consists of two main phases: \textit{\textbf{Phase~1}: Noise Transition Matrix Construction} and \textit{\textbf{Phase~2}: Corrupted Label Correction}. 

\textit{\textbf{Phase 1}} aims to \textit{capture the noise characteristics/structure inherent in the dataset}. 
\tool employs the Asymmetric Co-Training strategy to distinguish likely clean and noisy samples, from which it estimates a \textit{noise transition matrix}. This matrix captures the probability that each clean label is flipped into each noisy one, providing an approximation of the underlying noise distribution. It serves as a foundation for training a noise-aware model in the subsequent phase.

In \textit{\textbf{Phase 2}}, \textit{a noise-aware model is trained by incorporating the noise transition matrix into its training objective}. 
This integration allows the model to adjust its learning according to the inferred noise distribution. This training strategy enhances robustness by decreasing unreliable supervision and promoting stable learning under noisy conditions.
Once the loss-stabilization trigger signals that training dynamics have reached a steady state, the model's predictions are blended with the original labels to form refined soft labels. 
This soft and cautious relabeling strategy stabilizes updates and avoids premature commitment to potentially incorrect predictions.

These two phases are executed iteratively. In each iteration, \tool re-estimates the transition matrix based on the current refined dataset and retrains the noise-aware model accordingly. 
As the quality of the relabeled data improves over iterations, the estimated noise distribution and the model becomes more reliable, enabling further refinement. The process terminates when label updates converge, and the final soft labels constitute the corrected dataset.

\section{Noise-Aware Framework for Correcting Corrupted Labels}
\label{sec:approach}

\subsection{Noise Transition Matrix Construction}
\label{sec:phase1}

This phase aims to construct a \textit{noise transition matrix} $T$, which captures the underlying noise structure of the dataset by modeling the probability that a clean label is corrupted into a noisy one.
However, since only the noisy dataset $\widetilde{D}$ is observable, directly estimating $T$ is non-trivial. To address this, we first identify a subset of samples whose labels are likely to be clean, denoted as $D_c$ ($D_c \subset \widetilde{D}$). This clean subset serves as a proxy for estimating the clean label distribution. By aligning the clean label distribution estimated from $D_c$ with the observed noisy label distribution in $\widetilde{D}$, we can derive approximate noise transition probabilities and construct a noise transition matrix $T$.

\subsubsection{Identifying Clean Subset}

To identify a reliable clean subset $D_c$, \tool employs the Asymmetric Co-Training strategy~\cite{act-tritraining2017, act-noisylabel2023}. Specifically, \tool maintains two models with complementary roles, \textit{a reliable model} $\mathcal{F}_{r}$, which focuses on learning \textit{reliable patterns} and a \textit{noisy model} $\mathcal{F}_{n}$, which aims to expose to the full potentially \textit{noisy patterns}. 
In this setup, $\mathcal{F}_{r}$ is trained exclusively on the evolving clean subset $D_c \subset \widetilde{D}$, while $\mathcal{F}_{n}$ is trained on the entire noisy dataset $\widetilde{D}$.
The collaborative interaction of these two models enables \tool to filter out clean samples and progressively expand $D_c$.
Algorithm~\ref{alg:act_training} shows the clean subset identification process of \tool via asymmetric co-training, which operates on $E$ epochs in three stages: warm-up, exploration, and exploitation. 

The \textit{warm-up stage} (lines 4-7) initializes both models, $\mathcal{F}_{r}$ and $\mathcal{F}_{n}$, by training them on the entire dataset $\widetilde{D}$ for $E_w$ epochs. This step enables each model to learn basic data patterns, establishing a stable starting point before any asymmetric updates occur. 

After the warm-up stage, the asymmetric co-training process (lines 11–31) is performed iteratively. At each epoch, $\mathcal{F}_{r}$ is updated using the evolving clean subset $D_c$ (line 29), while $\mathcal{F}_{n}$ continues to be trained on the entire dataset $\widetilde{D}$ (line 30). Throughout this process, $D_c$ is progressively expanded using two distinct strategies: an \textit{exploration} strategy in the earlier epochs to maximize sample coverage, and an \textit{exploitation} strategy in the later epochs to enhance the quality of the selected samples.

The \textit{exploration stage} (lines 15-20) aims to initialize and expand the clean subset $D_c$ during the early training epochs. This stage prioritizes increasing sample coverage while maintaining a reasonable level of reliability. In this stage, \textit{\tool leverages the predictions of the noisy model $\mathcal{F}_{n}$ to identify likely clean samples, i.e., samples whose predicted labels match their original labels} (line 17).

This stage relies on the noisy model $\mathcal{F}_{n}$ rather than the reliable model $\mathcal{F}_{r}$ for two main reasons.
First, because $\mathcal{F}_{n}$ is trained on the full dataset, it is able to capture broader and more generalizable patterns, whereas  $\mathcal{F}_{r}$ is initially restricted to a small and potentially biased clean subset $D_c$. As a result, leveraging $\mathcal{F}_{n}$'s predictions \textit{during the early epochs} allows \tool to grow $D_c$ more effectively without being bottlenecked by its initial size.
Second, consistent with the well-established learning dynamics of DNNs, $\mathcal{F}_{n}$ tends to learn clean patterns early in the training before overfitting to noisy labels~\cite{arpit2017closer}.
This learning behavior makes $\mathcal{F}_{n}$ a reasonably reliable proxy for clean-label estimation at this early stage.
These factors enable \tool to rapidly construct a diverse and reasonably trustworthy clean subset $D_c$.

The \textit{exploitation stage} (lines 21-26) shifts the selection objective from maximizing coverage to enhancing the quality of the clean subset $D_c$. Once $D_c$ becomes sufficiently large to support reliable learning, aggressive expansion is no longer necessary and may introduce additional noise. Accordingly, \tool adopts a stricter selection criterion, \textit{only samples for which both models, $\mathcal{F}_{r}$ and $\mathcal{F}_{n}$, predict the same label as the original label are added to $D_c$} (line 23). This consensus-based filtering strategy reduces reliance on a single model's potentially noisy predictions and minimizes the risk of incorporating mislabeled samples. By enforcing agreement between a model trained on broad data patterns and one trained on the clean subset, \tool selects samples that are consistently supported by both perspectives, thereby resulting in a more reliable clean subset.

\begin{algorithm}[t]
\DontPrintSemicolon
\SetAlgoLined
\SetKwInOut{Input}{Input}
\SetKwInOut{Output}{Output}

\Input{
    Noisy dataset $\widetilde{D} = \{(x_i, \tilde{y}_i)\}_{i=1}^N$; \\
    Total number of training epochs $E$; \\
    Warm-up epochs $E_w$;
}
\Output{Clean subset $D_c$}

\BlankLine
\textbf{// Initialize clean subset} \\
$D_c \gets \emptyset$;

\BlankLine
\textbf{// Warm-up stage} \\
\For{$e \gets 1$ \KwTo $E_w$}{
    Train $\mathcal{F}_{r}$ on $\widetilde{D}$; \\
    Train $\mathcal{F}_{n}$ on $\widetilde{D}$;
}

\BlankLine
\textbf{//Switch point from exploration to exploitation} \\
$E_s \gets E_w + \left\lfloor \frac{E - E_w}{2} \right\rfloor$;

\BlankLine
\textbf{// Asymmetric Co-Training} \\
\For{$e \gets E_w + 1$ \KwTo $E$}{
    
    \BlankLine
    \For{$(x_i, \tilde{y}_i) \in \widetilde{D}$}{
        $y^{n}_i \gets \mathcal{F}_{n}(x_i)$; \\
        $y^{r}_i \gets \mathcal{F}_{r}(x_i)$; \\
         \BlankLine
        \If{$e \leq E_s$}{  % 
            \BlankLine
            \textbf{// Exploration stage} \\
            \If{$y^{n}_i = \tilde{y}_i$}{
                $D_c \gets D_c \cup \{(x_i, \tilde{y}_i)\}$;
            }
        }
         \BlankLine
        \Else{ 
            \BlankLine
            \textbf{// Exploitation stage} \\
            \If{$y^{n}_i = y^{r}_i$ \textbf{and} $y^{n}_i = \tilde{y}_i$}{
                $D_c \gets D_c \cup \{(x_i, \tilde{y}_i)\}$;
            }
        }
    }

    \BlankLine
    \textbf{// Asymmetric model updates} \\
    Train $\mathcal{F}_{n}$ on $\widetilde{D}$; \\
    Train $\mathcal{F}_{r}$ on $D_c$;
}

\BlankLine
\KwRet $D_c$;

\caption{Clean Subset Identification}
\label{alg:act_training}
\end{algorithm}

\subsubsection{Constructing Noise Transition Matrix}

This work models the noise characteristics of the noisy dataset $\widetilde{D}$ by constructing a noise transition matrix $T \in \mathbb{R}^{C \times C}$, where $C$ is the number of classes. 
The matrix $T$ summarizes how label corruption occurs from clean classes to noisy classes.
Each entry $T_{c,c'} = P(\tilde{y} = c' \mid y = c)$ represents the probability that a sample with clean label $y = c$ is observed as a noisy label $\tilde{y} = c'$.

Ideally, if ground-truth clean labels were available, the transition probability $T_{c,c'}$ could be computed directly by the following formula:

\begin{equation}
\label{eq:transition_with_groundtruth}
T_{c, c'} = \frac{\sum_{i=1}^{N} \mathbb{1}[y_i = c] \cdot \mathbb{1}[\tilde{y}_i = c']}{\sum_{i=1}^{N} \mathbb{1}[y_i = c]}
\end{equation}
where $\mathbb{1}[.]$ is the indicator function, returning 1 if its condition is true, e.g., $\mathbb{1}[y_i = c]$ returns 1 if $y_i = c$, and 0 otherwise.
However, the ground-truth clean labels are unavailable, we have access to noisy label $\tilde{y}$ only. Consequently, Eq.~\ref{eq:transition_with_groundtruth} cannot be directly applied.

In this work, to approximate the clean label distribution $P(y = c \mid x)$, \tool employs the reliable model $\mathcal{F}_{r}$, trained on the clean subset $D_c$. 
Accordingly, the transition probabilities can be estimated as:

\begin{equation}
\label{eq:transition_without_groundtruth}
T_{c, c'} = \frac{\sum_{i=1}^{N} P(y_i = c| x_i) \cdot \mathbb{1}[\tilde{y}_i = c']}{\sum_{i=1}^{N} P(y_i = c| x_i)}
\end{equation}
where $P(y_i = c \mid x_i) = \mathcal{F}_r(x_i)$ denotes the predicted probability that input $x_i$ belongs to class $c$.

The use of $\mathcal{F}_r$ to the estimate the clean label distribution is justified by its training on a clean subset, $D_c$. This enables $\mathcal{F}_r$ to learn reliable patterns for approximating the true label distribution. Moreover, by using soft probabilistic predictions rather than hard decisions, this approach mitigates the risk of introducing sharp errors into the estimation of the transition matrix.
Overall, Eq.~\ref{eq:transition_without_groundtruth} provides a practical solution for constructing the noise transition matrix in real-world scenarios where ground-truth clean labels are unavailable.

\subsection{Corrupted Label Correction}
The goal of this phase is to train a reliable model whose predictions can be used to guide the correction of corrupted labels.
In \tool, we incorporate the noise transition matrix $T$ into the training objective, making the model explicitly aware of the underlying noise label distribution. This integration encourages the model to down-weight unreliable supervision signals and emphasize trustworthy patterns. As a result, the model can be more robust to noisy labels and achieves better generalization performance. Once trained, the model's predictions are used as soft-label estimates that are blended with the observed labels for label correction. This soft correction strategy enables \tool to progressively refine noisy labels in a principled and stable manner. 

\subsubsection{Training a Noise-Aware Model}

In Phase~2, rather than initializing a new model from scratch, \tool leverages the reliable model $\mathcal{F}_r$ obtained from the previous phase as the starting point for subsequent training. Since $\mathcal{F}_r$  was trained on a highly clean subset $D_c$, it offers a stable initialization with reliable decision boundaries. In this phase, $\mathcal{F}_r$ is further trained on the entire dataset $\widetilde{D}$, enabling it to generalize beyond the clean subset while retaining its initial robustness.

The core idea of this phase is to train $\mathcal{F}_r$ as a noise-aware model that explicitly accounts for the noise distribution. Instead of fitting directly to  original noisy labels, the model is trained to predict clean labels, which are then projected through the noise transition matrix to reconstruct the noisy labels. The training loss is computed between the reconstructed noisy labels and the original noisy labels.

This design offers several benefits. First, by measuring the training loss on noise-reconstructed surrogates rather than the raw predictions, it mitigates the risk of overfitting to corrupted labels. Second, incorporating the noise transition matrix makes the model explicitly aware of the noise structure, allowing it to suppress unreliable gradients and focus on learning from the clean signals. Third, although $\mathcal{F}_r$ is trained on the entire noisy dataset $\widetilde{D}$, it is optimized to predict clean labels, making its outputs directly usable for label correction.

Specifically, at each training step, $\mathcal{F}_r$ predicts a clean label distribution $\mathbf{p_i} \in \mathbb{R}^C$ for each instance $x_i \in \widetilde{D}$,
where each entry  $\mathbf{p_i}[c] = P(y_i = c \mid x_i )$ presents the probability of class $c$ being the clean label of $x_i$.
To re-construct the \textit{noisy label distribution} $\mathbf{\hat{p}_i} \in \mathbb{R}^C$ given the model's belief about the clean label $\mathbf{p_i}$, \tool projects $\mathbf{p_i}$ through the noise transition matrix $T \in \mathbb{R}^{C \times C}$.

\begin{equation}
\mathbf{\hat{p}_i}[c'] = \sum_{c=1}^C T_{c,c'} \mathbf{p_i}[c] = (T^{\top} \mathbf{p_i})[c']
\end{equation}

The \textit{estimated noise distribution} $\mathbf{\hat{p}_i}$ is then compared with the observed noisy label distribution  $\mathbf{\tilde{p}_i}$ (Sec.~\ref{sec:correcting_corrupted_labels}) to measure the loss for training $\mathcal{F}_r$. In this work, to measure discrepancy of two distributions $\mathbf{\hat{p}_i}$ and $\mathbf{\tilde{p}_i}$,  \tool employs Kullback-Leibler (KL) divergence~\cite{kullback1951information} as the loss function. The impact of different loss functions is empirically evaluated in Sec.~\ref{sec:impact_loss_function}.

\begin{equation}
\mathcal{L} = \frac{1}{N}\sum_{i=1}^N KL(\mathbf{\tilde{p}_i} \parallel \mathbf{\hat{p}_i}) = \frac{1}{N}\sum_{i=1}^N \sum_{c'=1}^C\mathbf{\tilde{p}_i}[c']\log\frac{\mathbf{\tilde{p}_i}[c']}{\mathbf{\hat{p}_i}[c']}
\end{equation}

% Since $\mathbf{\tilde{p}_i}$ is one-hot, this can be simplified to:
% $$\mathcal{L} = - \frac{1}{N}\sum_{i=1}^N \log \mathbf{{\hat{p}_i}}[c'] $$

% Intuitively, this training scheme encourages the model to explain the observed noisy labels through its predictions of clean labels, filtered by a learned corruption process. As training progresses, the model learns to separate signal from noise, gradually improving its capacity to generalize across the full dataset. This phase runs for at most $E$ epochs, with early stopping applied when the loss converges.

\subsubsection{Refining Corrupted Labels}
\label{sec:correcting_corrupted_labels}
$\mathcal{F}_r$ is trained for $E$ epochs or until the training loss stabilizes, its predictions are used to refine corrupted labels.
At correction iteration~\footnote{Each correction iteration involves executing both phases of \tool.} $t$, let $\mathbf{\tilde{p}_i}^{(t-1)}$ denote the observed label distribution for instance $x_i$, which is obtained from the previous correction iteration, $t-1$.
For the initial iteration ($t= 0$), this corresponds to the original noisy label encoded as one-hot vector $\mathbf{\tilde{p}_i}^{(0)} \in \mathbb{R}^C$:

\begin{equation}
\label{eq:original_labels}
    \mathbf{\tilde{p}_i}^{(0)}[c'] = 
\begin{cases}
1, & \text{if } c' = \tilde{y}_i, \\
0, & \text{otherwise}.
\end{cases}
\end{equation}

After each correction iteration, the label distribution is updated to a new soft label $\mathbf{\tilde{p}_i}^{(t)}$ by blending current observed distribution $\mathbf{\tilde{p}_i}^{(t-1)}$ with the predictions $\mathbf{p_i}$ of $\mathcal{F}_r$.

\begin{equation}
\label{eq:label_update}
    \mathbf{\tilde{p}_i}^{(t)} = \alpha \cdot \mathbf{p_i} + (1 - \alpha) \cdot \mathbf{\tilde{p}_i}^{(t-1)}, \text{with} \; \alpha \in (0, 1)
\end{equation}

This soft update strategy balances the model's predictive confidence  with prior label estimates, preserving uncertainty and preventing abrupt corrections. As iterations proceed, corrupted labels are gradually smoothed toward stable, consistent distributions inferred from the model's predictions.

\subsection{Iterative Correction}

The corrupted label correction process, including both phases of \tool, is repeated for up to $\tau$ iterations or until the refined label is converged. For each sample $x_i$, \tool begins with its original label $\mathbf{\tilde{p}_i}^{(0)}$. At each iteration $t < \tau$, the label distribution is progressively refined from the previous estimate $\tilde{\mathbf{p}}_i^{(t-1)}$ to a more accurate version $\tilde{\mathbf{p}}_i^{(t)}$, ideally approaching the true clean label distribution.
Once the iterative correction process concludes, the final repaired dataset is produced by converting the soft label distribution into hard labels via the maximum a posterior estimate:

\begin{equation}
y_i^* = \arg\max_c \tilde{\mathbf{p}}_i^{(\tau)}[c]
\end{equation}
\section{Evaluation Methodology}
\label{sec:eval-method}
To evaluate the effectiveness of \tool in correcting corrupted labels, we seek to answer the following research questions (RQs):
\begin{itemize}
    \item \textbf{RQ1. Label Correction Effectiveness:} How
    accurate is \tool in correcting corrupted labels? 
    How does \tool's performance compare with SOTA label correction methods~\cite{simifeat,sidyp, selc}? 
    \item \textbf{RQ2. Downstream Utility:} To what extent do datasets corrected by \tool improve downstream model performance compared with (i) the original noisy datasets, (ii) datasets corrected by other label correction methods, and (iii) model-centric noise-robust training approaches?

    \item \textbf{RQ3. Intrinsic Analysis:} 
    How do the key components of \tool contribute to its performance? 
    
     \item \textbf{RQ4. Sensitivity Analysis:} How sensitive is \tool to different factors, such as hyperparameters and dataset size?

    \item \textbf{RQ5. Time Complexity Analysis:} What is the computational cost of \tool?
    
\end{itemize}

\subsection{Dataset}
\label{sec:dataset}

We evaluate \tool under two practical forms of label noise: (i) synthetic-but-realistic noise introduced by automated labeling techniques, and (ii) naturally occurring noise in real-world datasets.

For \textit{noise introduced by automated labeling techniques}~\cite{alchemist, desmond2021semi, zhu2024apt, zhu2024coral, wang2021want, tomanek2009semi}: We employ four representative labeling strategies that cover the major data annotation paradigms in the literature. These strategies are used to annotate five widely adopted datasets covering both image and text classification tasks.
For image datasets, we use \fashion~\cite{fashion-mnist},
\organ~\cite{medmnistv1, medmnistv2}, and
\resisc~\cite{resisc45}. For text datasets, we use 
\agnews~\cite{agnews-yahoo} and \yahoo~\cite{agnews-yahoo}. 
Each dataset $D$ is split into two disjoint subsets, 
$D = D_u \cup D_t$, 
where $D_u$ is the unlabeled portion to be annotated by automated labeling techniques, and then used for evaluating corrupted label correction methods. The remaining set $D_t$ contains clean labels and is reserved for evaluating downstream task performance.

The employed automated labeling techniques include:

\begin{itemize}
    \item \textbf{LLM-based Annotation (\textit{LLM})}: This approach leverages the general reasoning capability of LLMs to infer labels directly from raw instances. Following prior work in LLM-based annotation~\cite{zhu2024apt, zhu2024coral, wang2021want}, we adopt zero-shot prompting (details are available on our website~\cite{website}) to generate a label for each instance. In this work, we employ Qwen2.5-7B-Instruct~\cite{qwen2.5_model} for text data and Qwen2-VL-7B-Instruct~\cite{qwen2vl_model} for image data. These model are chosen for their moderate size, strong reasoning ability, and competitive performance on recent benchmarks~\cite{qwen2.5, bai2025qwen2}.
    
    \item \textbf{Supervised Learning (\textit{Superv.})}: This strategy involves training a classifier on a labeled set, and then applying it to predict labels for the remaining unlabeled data. Following common practice under limited supervision, we train an XGBoost classifier using a small set of labeled instances, then use it to annotate unlabeled data in $D_u$. To provide feature inputs for the classifier, we use BERT~\cite{bert} for text embeddings and CLIP~\cite{clip} for image embeddings.

    \item \textbf{Semi-Supervised Learning (\textit{Semi-Superv.})}: This approach~\cite{desmond2021semi} propagates labels from a labeled set to unlabeled instances based on similarity or distance metrics. In this work, we adopt the Label Spreading algorithm~\cite{label_spreading}, a well-established graph-based method, to propagate labels from a small set of labeled data to unlabeled data $D_u$.

    \item \textbf{Weak-Supervised Learning (\textit{Weak-Superv.})}: This technique produces labels by aggregating pseudo-labels obtained from multiple programmatic labeling functions.
    We adopt Alchemist~\cite{alchemist}, a SOTA weak supervision framework, to annotate unlabeled data in $D_u$.
    Specifically, label functions are automatically generated using Qwen2.5-7B-Instruct~\cite{qwen2.5_model}, and  Snorkel~\cite{snorkel} is employed to aggregate weak labels into final annotations.  Note that, this approach is applied only to text datasets, due to the lack of standardized programmatic labeling functions for image data.

\end{itemize}

For \textit{real-world label noise}, we include \clothing dataset~\cite{clothing1M} to further evaluate the robustness of corrupted label correction approaches in naturally occurring noise. \clothing is a large-scale image dataset containing 14 clothing categories with approximately 37.5K manually verified labels.  This dataset is widely used for studying real-world label noise in related studies~\cite{cola, wang2019symmetric, feng2024clipcleaner}.

A summary of datasets used in our evaluation is provided in Table~\ref{tab:dataset}. The \textit{Noisy Label Set} corresponds to $D_u$, which is labeled by either automated data labeling techniques or affected naturally occurring noise (i.e., \clothing), The \textit{Test Set} corresponds to $D_t$, containing clean labels, which are used for evaluating downstream model performance.

\begin{table}\centering
\caption{Dataset Overview}
\label{tab:dataset}
\resizebox{\columnwidth}{!}
{
\begin{tabular}{l|l|r |r |r }\toprule
\multirow{2}{*}{Dataset} &\multirow{2}{*}{Data Type} &\multirow{2}{*}{\#Labels} &\multicolumn{2}{c}{Dataset Size} \\\cmidrule{4-5}
& & & \begin{tabular}[c]{@{}c@{}}Noisy Label\\ Set\end{tabular} &\begin{tabular}[c]{@{}c@{}}Test\\ Set\end{tabular} \\\midrule
\fashion &image &10  & 10K &2.5K \\
\organ &image &11 &17.8K &4.5K \\
\resisc &image &45 &18.9K &4.7K \\
\agnews &text &4 &12K &3K \\
\yahoo &text &10&10K &2.5K \\
\clothing &image &14 &32.1K &5.4K \\
\bottomrule
\end{tabular}
}
\end{table}

\subsection{Evaluation Procedure}
\textbf{RQ1. Label Correction Effectiveness:} 
To assess the effectiveness of \tool in correcting corrupted labels, we compare its performance against three SOTA label correction methods, including \semi~\cite{simifeat, docta}, \selc~\cite{selc}, and \sidyp~\cite{sidyp}.
All methods are evaluated based on the \textit{error rate} of the dataset after correction, which reflects the proportion of mislabeled instances that remain after the correction process. This metric provides a direct measure of how accurately each method identifies and repairs corrupted labels.
The baseline methods operate as follows:

\begin{itemize}
    \item \semi~\cite{simifeat, docta} detects noisy instances by examining their neighboring instances' labels. The instances whose labels disagree with the majority of their neighbors are flagged as mislabeled and relabeled based on either \textit{voting} (\semiv) or \textit{ranking} (\semir) strategies. We report the results for both variants.
    \item \selc~\cite{selc} exploits the memorization dynamics of DNNs~\cite{arpit2017closer}. It first estimates a \textit{turning point} during training steps, before which the model predictions are assumed to be more reliable. \selc then uses the model's predictions prior to this turning point to gradually correct noisy labels.
    \item \sidyp~\cite{sidyp} generates candidate clean labels based on label information from neighboring instances. It then applies a diffusion-based denoising process to refine these candidates and produce corrected labels.
\end{itemize}

To ensure a fair comparison, all methods, including \tool and the baselines, use the same embedding models to encode feature representations, BERT~\cite{bert} for text data and CLIP~\cite{clip} for image data.

\textbf{RQ2. Downstream Utility:}
To evaluate the practical utility of the corrected datasets, we evaluate how effectively they support downstream model training. For each label correction method, we use its corresponding corrected dataset to train a classification model and then measure its performance on a clean test set. This experiment enables us to examine whether improving label quality can lead to better generalization in actual learning scenarios.

Furthermore, we also benchmark against \textit{model-centric} approaches, which aim to train noise-robust models directly on the noisy datasets without explicitly repairing labels.
This comparison enables us to assess whether cleaning the dataset using \tool can yields greater downstream benefits than relying solely on noise-robust learning algorithms.
Specifically, we consider the following model-centric baselines:

\begin{itemize}
    \item \dividemix~\cite{dividemix}: A semi-supervised learning framework designed for training models with label noise. It trains two networks simultaneously and models per-sample loss distribution to separate the training samples into a clean and noisy sets. Semi-supervised learning techniques are then applied to guess and refine labels of the noisy samples.
    
    \item \coteaching~\cite{han2018co}: A co-training strategy where two DNNs are trained simultaneously. In each mini-batch, each network selects a subset of small-loss samples (likely to be clean) and passes them to the other network for parameter updates. This design helps to mitigate the influence of noisy labels.
 
     \item \sceloss~\cite{sceloss}: 
     This method introduces the Symmetric Cross Entropy (SCE) loss, which combines standard Cross Entropy $\text{CE}(q, p)$ and its reverse form $\text{CE}(p, q)$, where $p$ denotes the ground-truth label distribution and $q$ denotes the model prediction. The resulting loss balances effective learning from clean labels with robustness to label noise.
     
    \item \sidyp~\cite{sidyp}: A denoising framework that leverages a Simplex Label Diffusion Model to refine noisy labels. Beyond label correction, the model trained under this framework is inherently robust to label noise and can be directly used for inference, following the original setup in their paper.
   
\end{itemize}

\textbf{RQ3. Intrinsic analysis:}
This experiment investigates how the main components and design choices of \tool contribute to its overall performance. 

For \textit{component analysis}, we conduct an ablation experiment to study the contribution of two phases in \tool's performance.
This experiment allows us to isolate the impact of Phase~1 alone and evaluate how much additional improvement in label quality is achieved by incorporating Phase~2.

For \textit{design choice analysis}, we examine how the key mechanisms designed in \tool affect its results. We analyze the importance of the \textit{soft label refinement strategy} by comparing \tool's performance with soft and hard label update strategies.
In the soft-update setting, labels are progressively refined using predicted probability distributions as defined in Eq.~\ref{eq:label_update}. Meanwhile, in the hard-update setting, the model predictions are converted to one-hot vectors via \texttt{argmax} operation, which are then used to overwrite the existing labels after each iteration. 
We also analyze the influence of \textit{loss function} variants used for training the noise-aware model in Phase~2. Specifically, we consider several widely used divergence-based objectives for comparing probability distributions, including L2 Loss~\cite{l2-loss}, Hellinger Distance~\cite{hellinger}, and KL Divergence~\cite{kullback1951information}.

\textbf{RQ4. Sensitivity Analysis:} We investigate how varying key factors such as hyperparameter configurations and dataset size impact \tool's effectiveness. Specifically, we examine the impact of several key hyperparameters, including the embedding backbone, the number of warm-up epochs, the number of correction iterations, and the blending coefficient $\alpha$ in Eq.~\ref{eq:label_update}. In addition, we systematically vary the size of the training dataset to assess how \tool's effectiveness scales with data availability.

\subsection{Evaluation Metrics}

To comprehensively evaluate the effectiveness of corrupted label correction methods, we adopt two complementary evaluation metrics: (i) \textit{Error Rate} and (ii) \textit{Downstream Model Performance}.

\textbf{\textit{Error Rate}} measures the proportion of incorrect labels that remain in the dataset after correction. This metric directly reflects the precision of the label repair process, where \textit{a lower error rate indicates more accurate label recovery}. Given a dataset $D = \{(x_i, y_i)\}_{i=1}^{N}$ with ground-truth labels $y_i$ for each instance $x_i$. Let  $D^* = \{(x_i, y_i^*)\}_{i=1}^{N}$ be the corrected dataset with $y_i^*$ denote the corrected label of $x_i$. The error rate of $D^*$ is computed as:

\begin{equation}
    \text{ErrorRate}(D^*) = \frac{1}{N} \sum_{i=1}^{N} \mathbb{1}[y_i \neq y_i^*].
\end{equation}

\textbf{\textit{Downstream Model Performance}} evaluates the practical utility of the corrected dataset $D^*$ for supporting downstream learning tasks. This is evaluated by training a classifier on $D^*$ and measuring its predictive performance on a clean test set. \textit{Higher performance indicates that the corrected labels provide more reliable supervision for model training.} 
In this work, we report the macro-averaged F1-Score (F1-Macro), which equally weights all classes and is robust under class imbalance. Let $C$ be the number of classes and $\text{F1}_c$ be the F1-score of class $c$. The F1-Macro is calculated as:

\begin{equation}
    \text{F1-Macro} = \frac{1}{C} \sum_{c=1}^{C} \text{F1}_c.
\end{equation}

\section{Experimental Results}
\label{sec:results}

\subsection{Label Correction Effectiveness}

\begin{table*}\centering
\caption{Dataset Error Rate (\%) before (Original) and  after applying label correction techniques}\label{tab:rq1_dataset_error_rate}
%\resizebox{ extwidth}{!}{ % use this if the table is too large
\begin{tabular}{l|l|rrrrrr}\toprule
Dataset & Labeling technique &Original &\semiv &\semir &\selc &\sidyp &\tool \\\midrule
\multirow{4}{*}{\agnews} &\llm &19.0 &22.1 &21.9 &19.0 &29.9 &\textbf{16.3} \\
&\superv &12.3 &11.6 &11.6 &12.0 &12.6 &\textbf{10.8} \\
&\semiSuperv &16.0 &12.8 &12.8 &15.3 &12.3 &\textbf{10.7} \\
&\weakSuperv &44.1 &33.6 &33.1 &34.5 &29.8 &\textbf{16.4} \\\midrule
\multirow{4}{*}{\yahoo} &\llm &40.0 &48.4 &48.3 &40.4 &41.2 &\textbf{35.9} \\
&\superv &47.1 &48.5 &48.4 &45.6 &45.1 &\textbf{42.0} \\
&\semiSuperv &55.9 &52.3 &52.2 &49.3 &49.6 &\textbf{42.1} \\
&\weakSuperv &78.3 &73.9 &74.0 &77.9 &72.8 &\textbf{68.0} \\\midrule
\multirow{3}{*}{\fashion} &\llm &30.1 &26.4 &26.2 &26.5 &\textbf{20.8} &21.8 \\
&\superv &12.0 &11.4 &11.4 &11.5 &13.6 &\textbf{10.7} \\
&\semiSuperv &14.6 &12.8 &12.8 &11.8 &14.0 &\textbf{10.6} \\\midrule
\multirow{3}{*}{\organ} &\llm &76.2 &74.8 &74.6 &74.5 &73.7 &\textbf{66.9} \\
&\superv &17.9 &\textbf{14.8} &\textbf{14.8} &16.1 &15.4 &\textbf{14.8} \\
&\semiSuperv &27.6 &27.4 &27.4 &25.9 &\textbf{21.5} &\textbf{21.5} \\\midrule
\multirow{3}{*}{\resisc} &\llm &35.9 &33.7 &33.5 &35.2 &29.7 &\textbf{21.2} \\
&\superv &14.7 &9.9 &9.8 &14.1 &10.3 &\textbf{9.2} \\
&\semiSuperv &11.6 &9.9 &9.9 &10.9 &10.4 &\textbf{8.7} \\\midrule
\multicolumn{2}{c|}{\clothing} &38.3 &35.5 &35.5 &34.9 &32.5 &\textbf{28.5} \\
\bottomrule
\end{tabular}
\end{table*}

Table~\ref{tab:rq1_dataset_error_rate} reports the dataset error rates initially produced by the automated labeling techniques (\textit{Original}) and the corresponding error rates after applying label correction methods. Overall, \textit{\tool consistently achieves the lowest error rates across all settings, demonstrating its superior capability in identifying and repairing corrupted labels}.

On average, applying \tool substantially reduces the original error rates by approximately 25\%.
For instance, under LLM-based annotation setting, where the average original error rate is about 40\%, \tool reduces this rate to around 32\%, yielding a relative reduction of about 20\%.
The improvement is even more pronounced under the most challenging setting, \textit{Weak-Superv.}, which exhibits an initial error rate of 61\%. In this case, \tool lowers this rate to 42\%, corresponding to a 31\% relative reduction.
These remarkable reductions demonstrate that \tool is robust and effective in diverse noise conditions, even under highly noisy labeling scenarios.

Compared to \semi, \tool consistently delivers superior label correction performance across all datasets and noise settings, achieving an average relative improvement of 19\% over both \semiv and \semir.
For example, on the \yahoo dataset, which exhibits extremely high noise levels (40--78\%), \tool improves the data quality by 15\%, reducing the noise rate to 36\%--68\%. 
Meanwhile, \semi struggles to provide a reliable correction for such a severe noise condition.
By leveraging neighboring information for detecting and correcting corrupted labels, \semi may inadvertently propagate incorrect labels rather than repair them. Under the LLM-based annotation setting, \semi even increases the error rate from 40\% to 48\%.
This highlights that relying solely on neighborhood consistency is insufficient when labels are heavily corrupted.

Similarly, \tool also demonstrates stronger and more stable performance than \selc, achieving relative improvements ranging from 10\% to 52\% across all experimental settings.  For example, on \agnews under the \weakSuperv configuration, \selc reduces the error rate from 44.1\% to 34.5\%, corresponding to a 23\% improvement.
However, this performance is still far weaker than that  of \tool, which achieves a substantially lower noise rate, i.e., 16.4\%. This corresponds to a 63\% improvement over the original dataset and a 52\% gain over \selc's result.

The performance gap can be attributed to fundamental limitations in \selc's design.
While \selc exploits the early-learning dynamics of DNNs to distinguish noisy from clean samples and use model predictions to guide label correction, its effectiveness heavily depends on accurately estimating the \textit{turning point} at which the model transitions from learning clean patterns to memorizing noise. In practice, identifying this turning point is non-trivial. Stopping too early often results in unstable models that underfit clean labels, whereas stopping too late leads to overfitting noisy annotations~\cite{bai2021understanding}.
Meanwhile, \tool does not rely on a predefined turning point. Instead, it explicitly trains a noise-aware model and performs label updates only when the model has reached a stable training state. This design mitigates overfitting to noise and enables \tool to achieve more robust performance across diverse noise conditions.

Although \sidyp obtains a slightly better performance than \tool in one specific case, \fashion labeled by \llm, where it yields a 4\% improvement, \tool surpasses \sidyp in all other settings, with gains of up to 46\%. For example, on  \clothing, \sidyp reduces the noise rate from 38.3\% to 32.4\% (a 16\% relative reduction), whereas \tool further lowers it to 28.5\%, achieving an additional 13\% improvement over \sidyp. 

Moreover, \sidyp sometimes degrades the data quality rather than improves it. For example, on \agnews labeled by \llm, the original noise rate is 19\%, yet applying \sidyp increases the error rate to 29.9\%. This suggests that its diffusion-based denoising mechanism can be sensitive to noise patterns and may fail under certain automated labeling scenarios.
In contrast, \tool maintains consistently strong and reliable performance across all datasets and labeling strategies. Its noise-aware model training and progressively soft label refinements enable \tool to adapt effectively to diverse noise characteristics and ensure more stable label correction outcomes.

\begin{figure}
    \centering
    \includegraphics[width=0.6\linewidth]{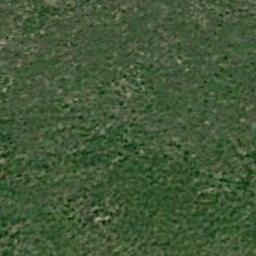}
    \caption{An image is correctly labeled as \texttt{meadow}}
    \label{fig:meadow}
\end{figure}

\begin{figure}
    \centering
    \includegraphics[width=0.6\linewidth]{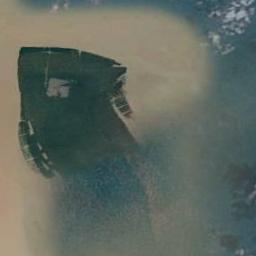}
    \caption{An image whose correct label is \texttt{island} but mislabeled as \texttt{cloud}}
    \label{fig:island}
\end{figure}

% %Examples
Figure~\ref{fig:meadow} presents an example from the \resisc dataset, where the instance is correctly labeled as \texttt{meadow} by the LLM-based annotator.
However, due to its visual appearance similar to a \texttt{forest}, all the studied label correction methods initially flagged this image as potentially mislabeled.
After correction, all the baselines, including \semi, \selc, and \sidyp, misleadingly relabeled this instance as \texttt{forest}.
In contrast, while \tool initially detected this instance as noisy, its two-phase framework, guided by a reliable noise-aware model, ultimately recovers the correct label \texttt{meadow}.

Figure~\ref{fig:island} shows another challenging case, an image whose true label is \texttt{island}, but incorrectly labeled as \texttt{cloud} by the LLM. Due to its ambiguous and visually fuzzy appearance, this example poses difficulties not only for automated labeling techniques but also for label noise detection and correction methods. The label correction methods exhibit varied behaviors in this case. Both variants of \semi relabeled the instance as \texttt{beach}, while \selc and \sidyp retained the incorrect label \texttt{cloud}. In contrast, \tool successfully identified the annotation error and accurately relabeled it as \texttt{island}, demonstrating \tool's robustness to ambiguous visual patterns.

\begin{figure}
    \centering
    \includegraphics[width=0.6\linewidth]{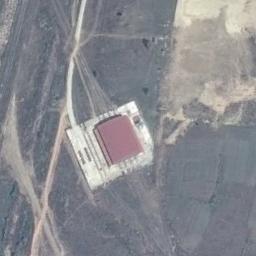}
    \caption{An image whose correct label is \texttt{railway station} but mislabeled as \texttt{stadium}}
    \label{fig:railway}
\end{figure}

However, \tool also faces limitations in cases where different categories share highly similar structural patterns. Figure~\ref{fig:railway} illustrates a case where the correct label is \texttt{railway station}, yet the instance is mislabeled as \texttt{stadium}. This image contains a large rectangular building with a prominent red roof. That visual characteristic is common across multiple categories, such as stadiums, sports complexes, stations, etc. In contrast, railway-specific signals are weak or absent. This strong semantic overlap makes the instance particularly challenging, and consequently, none of the label correction methods succeeds in repairing the label.
Specifically,  \semiv, \semir, \sidyp relabeled the instance as \texttt{airport}, while \tool and \selc preserved its original label \texttt{stadium}.
This example highlights an inherent limitation of label correction methods when the input features themselves are intrinsically ambiguous, making it difficult to obtain reliable predictions even with robust noise modeling. \newline\newline

\begin{gtheorem}
\textbf{Answer to RQ1}: \tool achieves the most accurate label correction across all datasets, noise types, and automated labeling strategies. On average, it reduces the original error rates by about 25\%, with even larger improvements under challenging settings such as \weakSuperv (achieving up to 31\% relative reduction). Compared to SOTA methods, \tool consistently outperforms \semi (by 19\% on average), \selc (by 10--52\%), and \sidyp (by up to 46\%). Moreover, unlike these baselines, \tool does not degrade data quality in any evaluation setting.
\end{gtheorem}

\subsection{Downstream Utility}

\subsubsection{Impact of Label Correction on Downstream Model Performance}

\begin{table*}\centering
\caption{Downstream model performance (F1-Macro) when training with dataset before (Original) and after label correction}\label{tab:rq2_downstream_performance}
%\resizebox{ extwidth}{!}{ % use this if the table is too large
\begin{tabular}{l|l|rrrrrr}\toprule
Dataset &Labeling technique  &Original &\semiv &\semir &\selc &\sidyp &\tool \\\midrule
\multirow{4}{*}{\agnews} &\llm  &78.62 &75.70 &75.83 &77.78 &61.69 &\textbf{82.33}  \\
&\superv  &88.11 &88.12 &88.12 &87.86 &86.56 &\textbf{88.86} \\
&\semiSuperv  & 87.19&86.82 &86.82 &86.51 &87.13 &\textbf{87.39} \\
&\weakSuperv   &58.71&65.93 &67.52 &65.69 &73.95 &\textbf{83.29}  \\\midrule

\multirow{4}{*}{\yahoo} &\llm  &58.55 &54.48 &54.04 &59.39 &58.10 &\textbf{60.01} \\
&\superv  &56.78 &53.70 &53.42 &56.70 &\textbf{56.77} &56.71 \\
&\semiSuperv & 51.62 &49.61 &49.60 &53.88 &52.74 &\textbf{56.33} \\
&\weakSuperv   &26.32 &27.54 &28.41 &23.87 &31.41 &\textbf{32.21} \\
\midrule
\multirow{3}{*}{\fashion} &\llm  &72.83 &72.72 &73.34 &73.84 &\textbf{78.89} &77.76 \\

&\superv  &88.97 &88.05 &88.61 &88.77 &84.65 &\textbf{88.76} \\
&\semiSuperv  &81.10 &87.74 &87.51 &88.23 &84.77 &\textbf{88.57}  \\\midrule
\multirow{3}{*}{\organ} &\llm &15.15  &15.56 &15.72 &15.77 &14.74 &\textbf{19.55} \\
&\superv  &86.83 &\textbf{87.10} &\textbf{87.10} &86.80 &85.17 &82.68 \\
&\semiSuperv &80.37  &80.05 &80.05 &\textbf{82.29} &80.05 &75.11 \\\midrule
\multirow{3}{*}{\resisc} &\llm  &62.52 &62.89 &62.96 &61.66 &65.54 &\textbf{77.16} \\
&\superv  &89.49 &90.86 &91.15 &89.91 &90.73 &\textbf{91.68} \\
&\semiSuperv &90.48  &90.47 &90.37 &91.57 &90.26 &\textbf{92.10} \\\midrule
%clothing
\multicolumn{2}{c|}{\clothing}  &61.03    &62.14 &62.10 &62.24 &63.71 &\textbf{65.68} \\

\bottomrule
\end{tabular}
\end{table*}

Table~\ref{tab:rq2_downstream_performance} shows the performance of downstream models trained on datasets before and after applying label correction techniques. As seen, \textit{\tool often yields the strongest downstream performance across datasets and labeling conditions}. By substantially improving label quality, \tool enables downstream models to achieve about a 10\% increase in F1-Macro compared to training on the original corrupted datasets. In addition, \tool also outperforms existing label correction methods by margins of up to 35\%.

Interestingly, the improvements provided by \tool are most pronounced under severe or structurally complex noise scenarios. For example, on the \agnews under \weakSuperv labeling, the original dataset contains 44.1\% mislabeled instances, resulting in a weak classifier with only 58.71 in F1-Macro. After correction, \tool boosts the downstream performance by 42\%, which surpasses the other baselines 13--27\%. Similarly, on the \resisc under LLM-based annotation, where 35.9\% of labels are incorrect, the downstream model trained on that corrupted data achieves only 62.52 in F1-Macro. With \tool's corrections, the model performance increases to 77.16, outperforming the baselines by about 22\%. 
\textit{These considerable gains demonstrate that \tool not only repairs incorrect labels effectively but also produces higher-quality training datasets that lead to more robust and generalizable models.}

While \tool is highly effective in most settings, we observe a few cases, such as  \organ under \superv or \semiSuperv labeling, where baselines like \semi or \selc achieve stronger downstream results. For instance, on the \organ with \superv labeling, both \semi and \tool can reduce the dataset error rate to 14.8\%. However, the downstream classifier trained on the dataset corrected by \semi reaches an F1-Macro of 87.10, which is approximately 5\% higher than the model trained on \tool's corrected dataset.
This result suggests that downstream utility depends not only on overall correction accuracy but also on which specific instances are corrected. Some corrected samples may provide more informative learning signals or better preserve label–feature relationships, thereby leading to improved model generalization.
In future work, we plan to enhance \tool by incorporating instance selection strategies and modeling label-feature interactions to further improve both correction effectiveness and the generalization capability of downstream models.

Note that, all experiments reported in Table~\ref{tab:rq2_downstream_performance} were conducted using the same standard MLP classifier with two hidden layers consisting of 512 and 256 units, respectively. To validate the generality of the observed trends, we also evaluated additional model architectures and found consistent performance improvements across settings. Full implementation details and extended experimental results are available on our project website~\cite{website}.

\subsubsection{Comparison with the Model-centric Approaches}

\begin{table*}\centering
\caption{Downstream model performance (F1-Macro) of models trained on datasets corrected by \tool compared with Model-centric approaches}
\label{tab:rq2_model_centric}
\begin{tabular}{l|l|rrrrr}\toprule
Dataset & Labelling technique & \sceloss & \coteaching & \dividemix & \sidyp & \tool \\\midrule
\multirow{2}{*}{\agnews} &\llm& 78.37 & 68.56 & 80.01 & 62.06 & \textbf{82.33} \\
&\superv &88.18 & 87.39 & 88.10 & 88.82 & \textbf{88.86} \\
\midrule
\multirow{2}{*}{\yahoo} &\llm & 57.72 & 59.92 & 57.84 & 59.70 & \textbf{60.01} \\
&\superv & 56.60 & 56.28 & \textbf{58.90} & 58.66 & 56.71 \\\midrule
\multirow{2}{*}{\fashion} &\llm & 72.55 & 74.81 & 76.80 & 75.55 & \textbf{77.76} \\
&\superv & 89.02 & 75.88 & 85.54 & \textbf{90.84} & 88.76 \\\midrule
\multirow{2}{*}{\organ} &\llm & 15.49 & 15.11 & 16.51 & 16.42 & \textbf{19.55} \\
&\superv & 86.47 & 66.94 & 85.33 & \textbf{90.16} & 82.68 \\\midrule
\multirow{2}{*}{\resisc} &\llm & 61.97 & 46.34 & 71.29 & 61.55 & \textbf{77.16} \\
&\superv & 89.81 & 77.46 & 92.60 & \textbf{92.80} & 91.68 \\\midrule
\multicolumn{2}{c|}{\clothing} &61.46 & 62.73 & 65.00 & 63.14 & \textbf{65.68} \\
\bottomrule
\end{tabular}
\end{table*}

Table~\ref{tab:rq2_model_centric} compares the downstream performance of models trained on datasets corrected by \tool against several noise-robust model-centric approaches. We report the results of two representative labeling conditions: (i) \textit{LLM}, which introduces the highest noise rates, and (ii) \textit{Superv.}, which produces the lowest noise rates in our study.
Results for the other labeling techniques are available on our website~\cite{website}.
To ensure a fair comparison, all the methods, including \tool, \sceloss, \coteaching, and \dividemix are trained using the same standard MLP architecture, while  \sidyp is evaluated using its original simplex diffusion model.

\textit{Under the \superv labeling setting, where noise level is relatively moderate (around 15\%), 
all methods produce reasonably strong downstream performance}. As expected, \textit{\sidyp, achieves the highest results}, benefiting from its sophisticated diffusion-based architecture. 
For example, on \fashion, \sidyp obtains F1-Macro of 92.84, slightly outperforming the model trained on \tool's corrected dataset (88.76). 
This result is reasonable, as advanced models like diffusion networks can effectively exploit the underlying clean structure when label noise is limited. Nevertheless, in this scenario, \tool remains highly competitive, despite relying on a lightweight MLP classifier.

\textit{When the noise becomes severe, as in an LLM-based labeling scenario, \tool consistently outperforms all model-centric baselines.} 
Under such extreme noise conditions, relying solely on robust training algorithms becomes insufficient, as the supervision signals are heavily contaminated.
For example, on the \resisc dataset labeled by \llm (35.9\% error rate), \tool enables the downstream classifier to reach an F1-Macro of 77.16, exceeding the performance of the other baselines by margins ranging from 8\% to 67\%. Notably, \tool outperforms \sidyp, which is built on a significantly more complex architecture, by 25\%.
These results indicate that explicit label repair play a crucial role, especially in high noise rate scenarios.

\begin{gtheorem}
\textbf{Answer to RQ2}: \tool significantly improves downstream model performance across all datasets and noise conditions by producing cleaner training data. On average, it yields 10\% improvement in F1-Macro compared to training on the original corrupted datasets and outperforms prior label correction methods by margins of up to 35\%.
Compared to noise-robust \textit{model-centric} approaches, \tool provides superior downstream accuracy, particularly under high noise condition. This demonstrates that cleaning the data is more beneficial than relying solely on robust training algorithms.
\end{gtheorem}

\subsection{Intrinsic Analysis}

\subsubsection{Component Analysis}

%Please add the following packages if necessary:
%\usepackage{booktabs, multirow} % for borders and merged ranges
%\usepackage{soul}% for underlines
%\usepackage{xcolor,colortbl} % for cell colors
%\usepackage{changepage,threeparttable} % for wide tables
%If the table is too wide, replace \begin{table}[!htp]...\end{table} with
%\begin{adjustwidth}{-2.5 cm}{-2.5 cm}\centering\begin{threeparttable}[!htb]...\end{threeparttable}\end{adjustwidth}
\begin{table}\centering
\caption{Contribution of two phases on \tool's performance}
\label{tab:rq3_component_analyisi}
\resizebox{\columnwidth}{!}{ % use this if the table is too large
\begin{tabular}{l|l|r|r|r}\toprule
& Dataset& Original &\begin{tabular}[c]{@{}c@{}}Phase 1 \\ Only \end{tabular} & \begin{tabular}[c]{@{}c@{}}Full\\ \end{tabular} \\\midrule
\multirow{2}{*}{\textbf{Error Rate (↓)}} &\resisc &35.90 &32.57 &21.23 \\
&\agnews & 19.00 &20.06 &16.25 \\\midrule
\multirow{2}{*}{\textbf{F1-Macro (↑)}} &\resisc & 62.52 &65.00 &77.38 \\
&\agnews &78.62&78.39 &83.18 \\
\bottomrule
\end{tabular}
}
\end{table}

Table~\ref{tab:rq3_component_analyisi} evaluates the contribution of \tool's two-phase design. The \textit{Phase~1 only} variant reports performance when label correction relies solely on the predictions of the reliable model $\mathcal{F}_r$ trained in Phase~1, while the \textit{Full} variant corresponds to the complete framework that integrates both phases. 
Note that Phase~2 cannot be evaluated in isolation, as the noise-aware model in Phase~2 is trained based on the outputs of Phase~1.

Overall, \tool obtains its best performance when both phases are incorporated. For example, on \resisc, \textit{Phase~1 only} can reduce the error rate from 35.90\% to 32.57\%, demonstrating that the reliable model $\mathcal{F}_r$ trained in Phase~1 can partially mitigate label noise.
By further applying Phase~2, the error rate is substantially reduced to 21.23\%, corresponding to a relative reduction of 41\% from the original dataset. This improvement in label quality also significantly enhances the downstream performance, with F1-Macro increasing from 62.52 to 77.38.

However, Phase~1 alone does not consistently improve label quality across all settings. For example, on the \agnews dataset, using only Phase~1  slightly increases the error rate (from 19\% to 20.06\%). 
This indicates that directly using model predictions without explicitly accounting for the noise structure can be insufficient and may even propagate residual noise.
Nevertheless, Phase~1 still provides a necessary foundation for Phase~2. By leveraging outputs of Phase~1 to explicitly train a noise-aware model and performing iterative refinement, the full \tool framework effectively corrects labels and reduces the error rate to 16.25\%, while also achieving the best downstream performance.

\textit{These results highlight the complementary roles of the two phases: Phase~1 establishes a reliable foundation, while Phase~2 is crucial for robust noise modeling and stable label refinement, ultimately leading to consistently improved label quality and downstream model performance.}
\subsubsection{Impact of Soft Label Refinement Strategy}

%Please add the following packages if necessary:
%\usepackage{booktabs, multirow} % for borders and merged ranges
%\usepackage{soul}% for underlines
%\usepackage{xcolor,colortbl} % for cell colors
%\usepackage{changepage,threeparttable} % for wide tables
%If the table is too wide, replace \begin{table}[!htp]...\end{table} with
%\begin{adjustwidth}{-2.5 cm}{-2.5 cm}\centering\begin{threeparttable}[!htb]...\end{threeparttable}\end{adjustwidth}
\begin{table}[]
\caption{Impact of soft label refinement strategy on \tool's performance}
\label{tab:rq3_soft_labels}
% \resizebox{\columnwidth}{!}{ 
\begin{tabular}{l|l|r|r|r}\toprule
& Dataset& Original &\begin{tabular}[c]{@{}c@{}}Hard \\ Labels \end{tabular} &\begin{tabular}[c]{@{}c@{}}Soft \\ Labels \end{tabular} \\\midrule
\multirow{2}{*}{\textbf{Error Rate (↓)}} &\resisc & 35.90 &28.72 &21.23 \\
&\agnews &19.00&17.22 &16.25 \\\midrule
\multirow{2}{*}{\textbf{F1-Macro (↑)}} &\resisc & 62.52&69.08 &77.38 \\
&\agnews &78.62&82.03 &83.18 \\
\bottomrule
\end{tabular}
% }
\end{table}

Table~\ref{tab:rq3_soft_labels} compares \tool's performance under soft and hard label update strategies.
As shown, soft label refinement consistently outperforms hard label updates, demonstrating the effectiveness of a progressive and stable correction process. On average, soft updates reduce dataset error rates by approximately 16\% and improve downstream model performance by up to 12\% compared to hard updates.
For example, on \resisc, hard label updates reduce the original error rate from 35.90\% to 28.72\%, while soft label refinement further lowers it to 21.23\%.
This substantial gap indicates that propagating full predictive distributions enables more accurate and robust label correction than committing to one-hot predictions at each iteration.
\textit{These results confirm that soft label refinement is a critical design choice in \tool. By preserving uncertainty and avoiding premature hard decisions, the soft update mechanism mitigates error propagation across iterations, leading to more accurate label repair and better downstream generalization.}
\subsubsection{Impact of Loss Function}
\label{sec:impact_loss_function}

%Please add the following packages if necessary:
%\usepackage{booktabs, multirow} % for borders and merged ranges
%\usepackage{soul}% for underlines
%\usepackage{xcolor,colortbl} % for cell colors
%\usepackage{changepage,threeparttable} % for wide tables
%If the table is too wide, replace \begin{table}[!htp]...\end{table} with
%\begin{adjustwidth}{-2.5 cm}{-2.5 cm}\centering\begin{threeparttable}[!htb]...\end{threeparttable}\end{adjustwidth}
\begin{table}\centering
\caption{Impact of loss functions on \tool's performance}\label{tab:rq3_loss_function}
\resizebox{\columnwidth}{!}{ % use this if the table is too large
\begin{tabular}{l|l|c|c}\toprule
Dataset&Loss variants &\textbf{Error Rate (↓)} &\textbf{F1-Macro (↑)} \\\midrule
\multirow{3}{*}{RESISC45} &L2 &24.95 &72.33 \\
&Hellinger &27.56 &69.59 \\
&KL Diverage &21.23 &77.38 \\\midrule
\multirow{3}{*}{AGNews} &L2 &16.58 &82.72 \\
&Hellinger &18.57 &80.01 \\
&KL Diverage &16.25 &83.18 \\
\bottomrule
\end{tabular}
}
\end{table}

Table~\ref{tab:rq3_loss_function} shows the performance of \tool under different loss functions used to optimize the noise-aware model $\mathcal{F}_r$ in Phase~2. Overall, \textit{KL Divergence yields the best results, achieving the lowest error rates and highest downstream F1-Macro scores}. For example, on \resisc, when Hellinger or L2 is used, the corrected datasets retain error rates of 27.56\% and 24.95\%, respectively. Meanwhile, with KL Divergence, the error rate is reduced to 21.23\%. This demonstrates that the choice of loss function has a substantial impact on the robustness of $\mathcal{F}_r$ and, consequently, on the overall label correction quality.

This result can be explained by the role and design of Phase~2 in \tool. In this phase, the supervision signals for training $\mathcal{F}_r$ are not hard labels, but \textit{soft supervision signals} derived from the estimated noise distributions and the refined label probabilities. 
In this setting, KL divergence is particularly well-suited, as it explicitly measures the discrepancy between two probability distributions and assigns strong penalties to confident but incorrect predictions. This property encourages the model to align its predictions with the refined label distributions, enabling more effective correction of corrupted labels.
Meanwhile, L2 is too insensitive, as it treats all prediction errors uniformly.  
Hellinger is distribution-aware, but it is too conservative since it smooths the difference through the square-root transformation and bounded range.
As a result, both L2 and Hellinger are less suitable for training $\mathcal{F}_r$ in \tool.

\begin{gtheorem}
\textbf{Answer to RQ3}: Two phases in \tool play complementary roles: Phase~1 establishes a reliable foundation for training the model in the subsequent phase, while Phase~2 is essential for robust noise modeling and stable label refinement. In addition, adopting KL divergence as the loss function and performing iterative soft-label updates collectively lead to improve label quality and superior downstream model performance.
\end{gtheorem}
\subsection{Sensitivity Analysis}

\subsubsection{Impact of Embedding Models}

%Please add the following packages if necessary:
%\usepackage{booktabs, multirow} % for borders and merged ranges
%\usepackage{soul}% for underlines
%\usepackage{xcolor,colortbl} % for cell colors
%\usepackage{changepage,threeparttable} % for wide tables
%If the table is too wide, replace \begin{table}[!htp]...\end{table} with
%\begin{adjustwidth}{-2.5 cm}{-2.5 cm}\centering\begin{threeparttable}[!htb]...\end{threeparttable}\end{adjustwidth}
\begin{table}\centering
\caption{Impact of embedding models on \tool's performance}\label{tab:rq3_embedding_models}
% \resizebox{ extwidth}{!}{ % use this if the table is too large
\begin{tabular}{l|l|c|c}\toprule
Dataset& Model&\textbf{Error Rate (↓)} &\textbf{F1-Macro (↑)} \\\midrule
\multirow{4}{*}{RESISC45} &ResNet50 &33.36 &61.45 \\
&Dinov3 &23.95 &73.90 \\
&SigLIP &22.85 &74.88 \\
&CLIP &21.23 &77.38 \\\midrule
\multirow{4}{*}{AGNews} &BGE-M3 &16.85 &82.68 \\
&XLNET &18.15 &79.50 \\
&RoBERTa &15.79 &83.34 \\
&BERT &16.25 &83.18 \\
\bottomrule
\end{tabular}
\end{table}

Table~\ref{tab:rq3_embedding_models} shows how the choice of embedding models affects \tool's performance. As expected, 
\textit{\tool benefits from stronger embeddings}, resulting in lower error rates and higher F1-Macro scores.
On the image dataset \resisc, \tool achieves the best performance when using CLIP~\cite{clip}, with 21.23\% error rates and 77.38 in F1-Macro. 
This result is consistent with prior studies~\cite{liu2025data}, which show that CLIP provides more semantically rich embeddings and generally outperforms alternative vision encoders such as DINO~\cite{simeoni2025dinov3} or SigLIP~\cite{zhai2023sigmoid}.

A similar trend is observed on the text dataset \agnews. 
Although the noise level in this dataset is relatively moderate, leading to smaller performance gaps across embedding models, stronger language encoders still yield better results. In particular, the more powerful model RoBERTa~\cite{liu2019roberta} outperforms the older models like XLNet~\cite{yang2019xlnet}. These results confirm that \textit{while \tool is compatible with a wide range of pretrained embedding models, its effectiveness is enhanced when leveraging high-quality representations}.

\subsubsection{Impact of the number of warm-up epochs}

\begin{figure}
\centering
\begin{subfigure}{\columnwidth}
\centering
\includegraphics[width=\columnwidth]{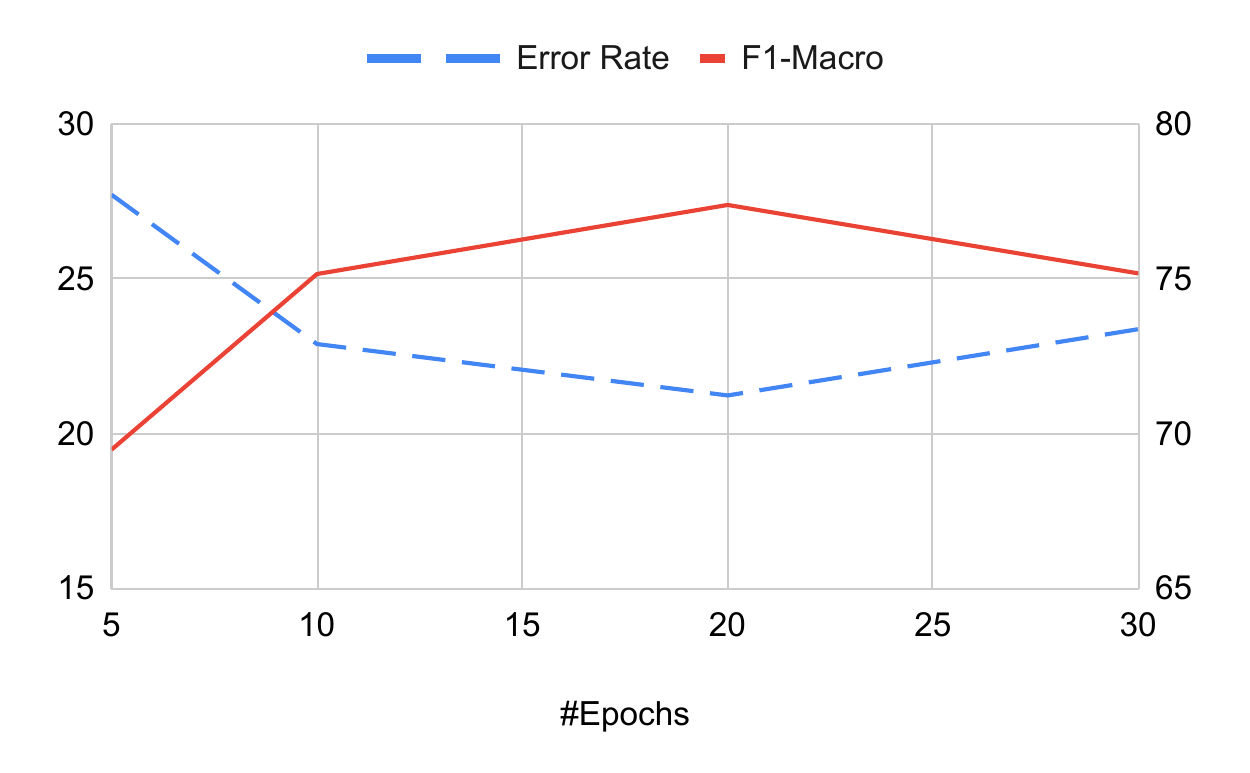}
\caption{\resisc}
\label{fig:resisc_warmup}
\end{subfigure}\\

\begin{subfigure}{\columnwidth}
\centering
\includegraphics[width=\columnwidth]{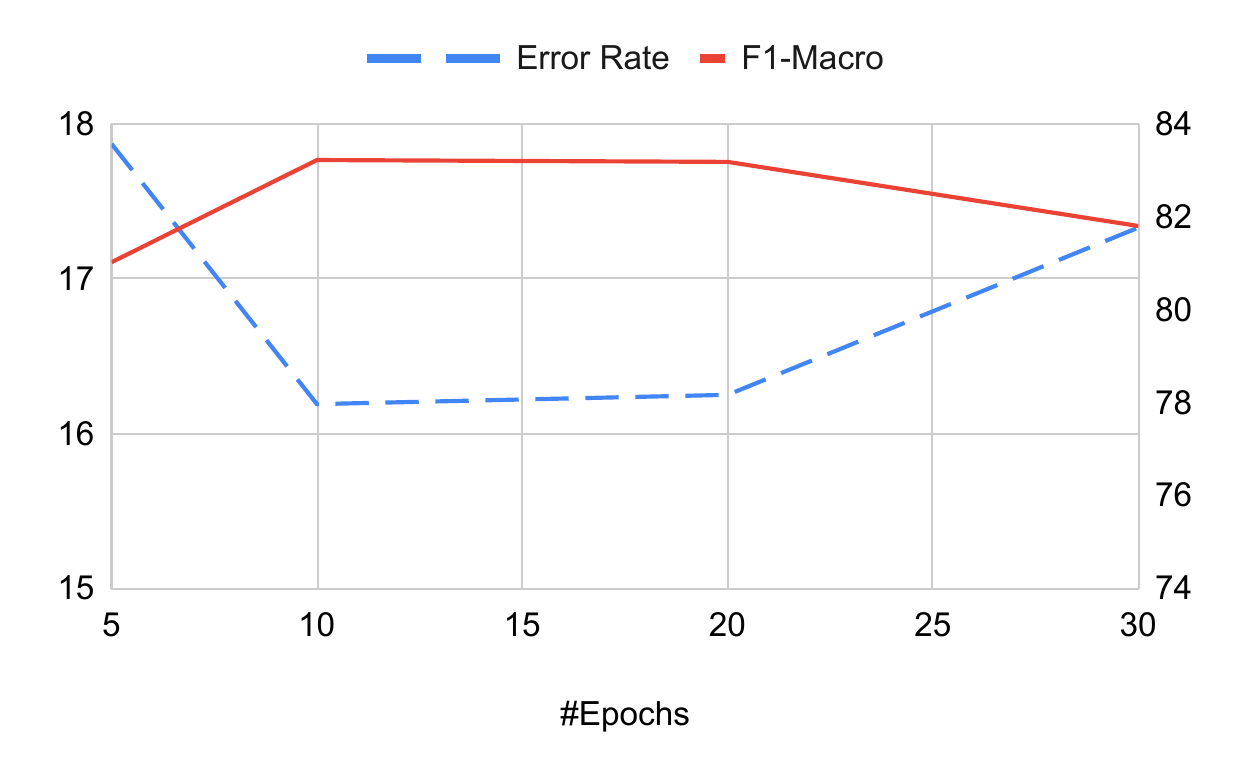}
\caption{\agnews}
\label{fig:agnews_warmup}
\end{subfigure}

\caption{Impact of the number of warm-up epochs ($E_w$ in Phase~1) on \tool's performance, left axis: \textit{Error Rate}; right axis: \textit{F1-Macro}}
\label{fig:warmup_epochs}
\end{figure}

In Phase~1, \tool initializes both models $\mathcal{F}_n$ and  $\mathcal{F}_r$ by training them on the entire noisy dataset $\widetilde{D}$ for $E_w$ warm-up epochs to acquire basic task knowledge. Figure~\ref{fig:warmup_epochs} shows the effect of varying $E_w$ on \tool's overall performance. The results indicate that \textit{\tool achieves its optimal performance when the warm-up stage is set to approximately 15--20 epochs.}
With fewer warm-up epochs, both $\mathcal{F}_n$ and $\mathcal{F}_r$ remain under-trained, leading to unstable predictions. In contrast, extending the warm-up stage beyond this range results in performance degradation.
This phenomenon aligns with the well-known learning dynamics of DNN under noise conditions~\cite{arpit2017closer}, where models initially capture clean patterns but gradually memorize noisy labels as training continues. Prolonged training on  $\widetilde{D}$ therefore increases the risk that $\mathcal{F}_r$ overfits to corrupted labels, weakening its reliability.

These results highlight a critical trade-off, \textit{the warm-up stage should be sufficiently long to establish useful representations, but short enough to avoid noise memorization}. 
Accordingly, after 15-20 warm-up epochs, \tool stops training $\mathcal{F}_r$ on $\widetilde{D}$ and instead updates its parameters using the evolving clean dataset $D_c$. This preserves robustness and enables more accurate noise modeling and label correction.
\subsubsection{Impact of the number of correction iterations}

\begin{figure}
    \centering
    \includegraphics[width=\linewidth]{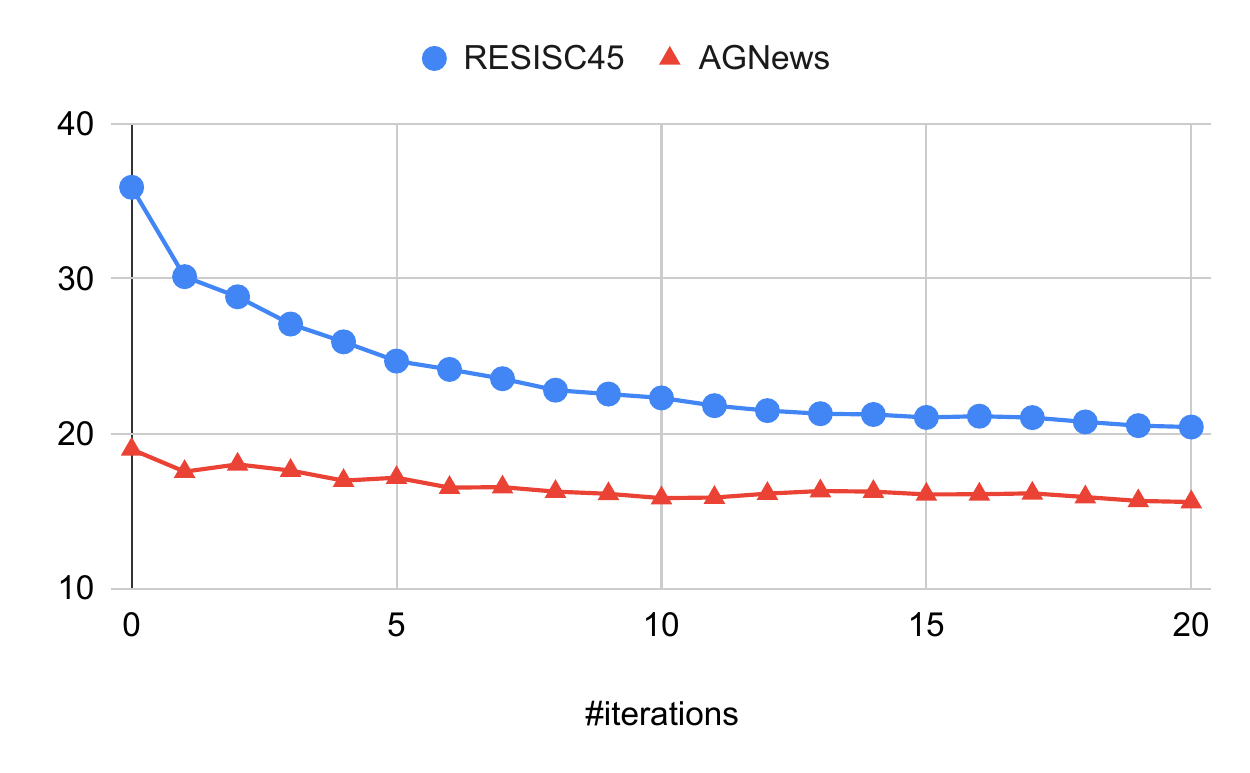}
    \caption{Impact of the number of correction iterations on \tool's performance, measured by \textit{Error Rate}}
    \label{fig:impact_iterations}
\end{figure}

Figure~\ref{fig:impact_iterations} demonstrates that \textit{increasing the number of correction iterations~\footnote{An iteration consisting of full run both Phase~1 and phase~2}} \textit{improves the overall repaired label quality.}  
In particular, with a single iteration, the error rate on \resisc decreases from 35.9\% to 30.1\%, while that on \agnews reduces from 19.0\% to 17.5\%. When the number of iterations  increases to 10, the error rates are further reduced to 22.3\% and 15.8\%, respectively.
However, beyond this point, the performance gains gradually saturate. This behavior suggests that \textit{\tool should run label correction for approximately 10 iterations to achieve a balance trade-off between correction effectiveness and computational cost.}

\subsubsection{Impact of the blending coefficient}

\begin{figure}
\centering
\begin{subfigure}{\columnwidth}
\centering
\includegraphics[width=\columnwidth]{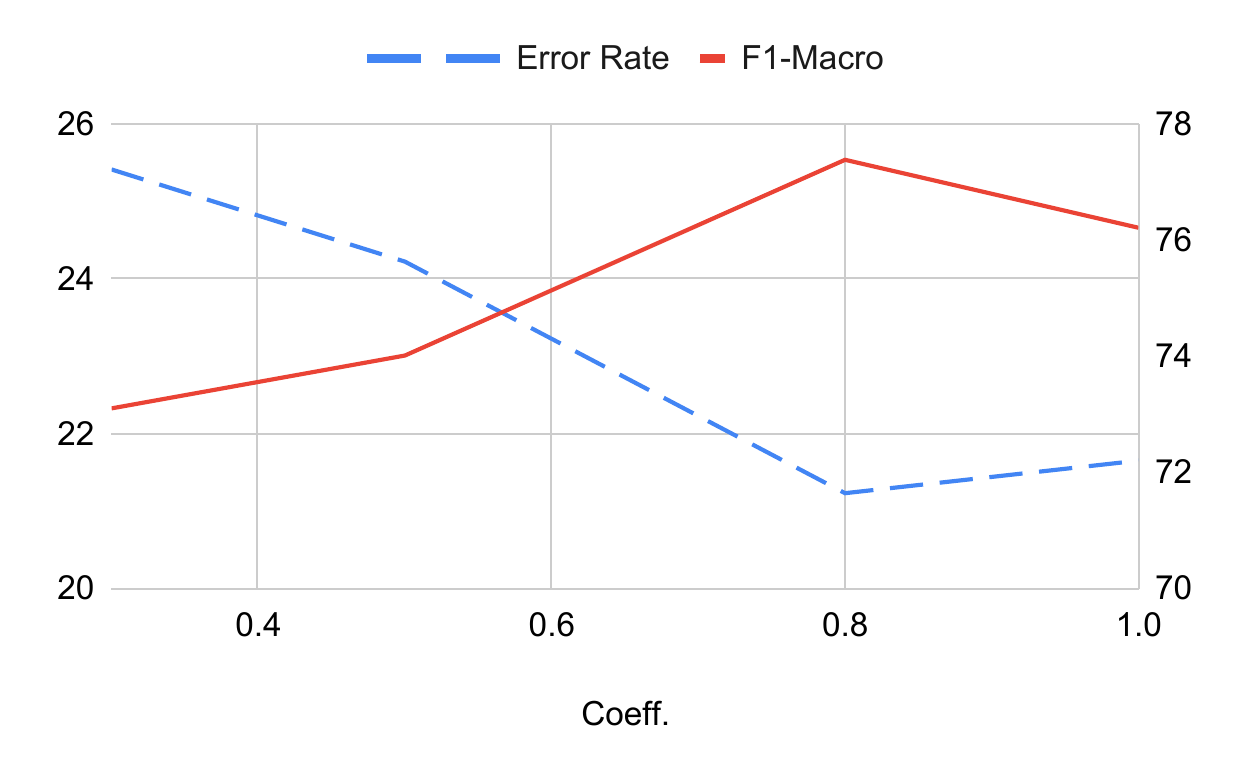}
\caption{\resisc}
\label{fig:resisc_alpha}
\end{subfigure}\\

\begin{subfigure}{\columnwidth}
\centering
\includegraphics[width=\columnwidth]{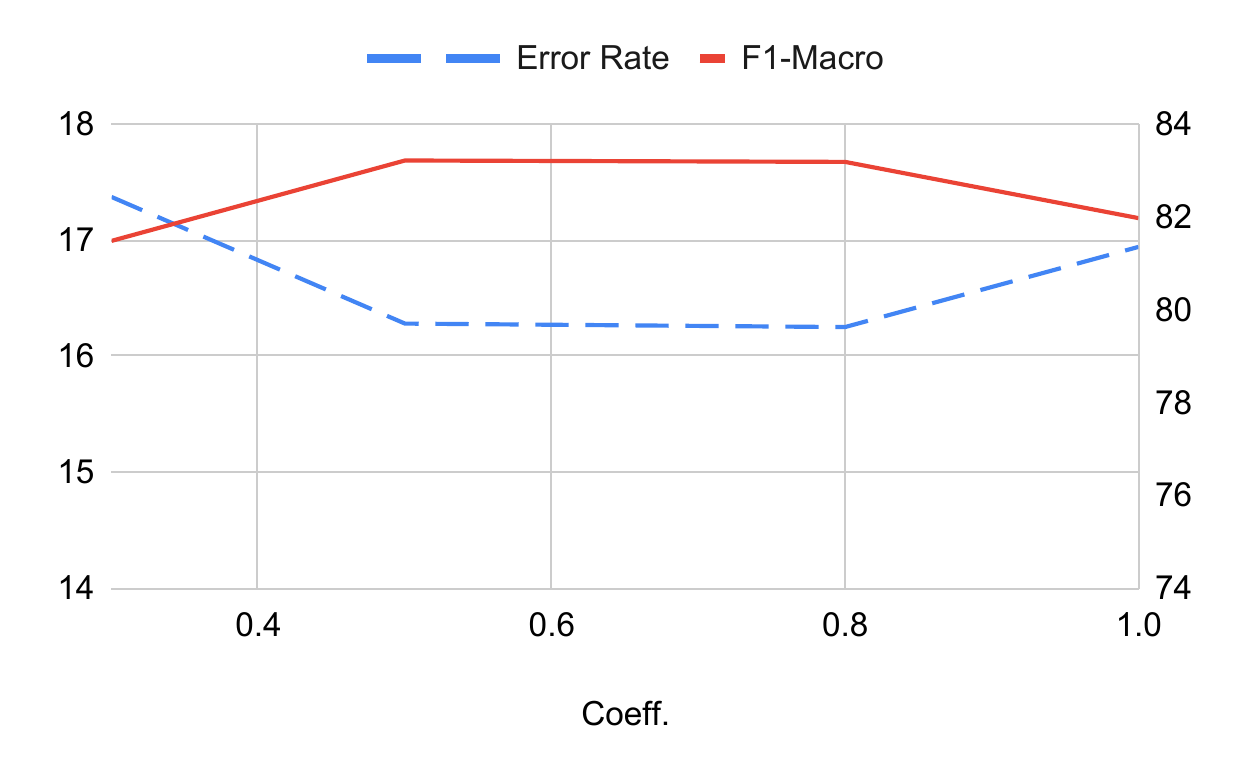}
\caption{\agnews}
\label{fig:agnews_alpha}
\end{subfigure}

\caption{Impact of the blending coefficient $\alpha$ (Eq.~\ref{eq:label_update}) on \tool's performance, left axis: \textit{Error Rate}; right axis: \textit{F1-Macro}}
\label{fig:alpha}
\end{figure}

Figure~\ref{fig:alpha} illustrates the impact of blending coefficient $\alpha$ (Eq.~\ref{eq:label_update}), which controls the trade-off between model's prediction $\mathbf{p}_i$ and the previously refined label distribution $\mathbf{\tilde{p}}_i^{(t-1)}$ during soft label updates. On the highly noisy dataset \resisc (35.9\% error rate), \tool obtains its optimal performance at a relatively large value of $\alpha$ (i.e., $\alpha = 0.8$). 
This indicates that under severe noise conditions, placing greater emphasis on the model's predictions is more reliable than historical label estimates, which may still preserve substantial noise.
In contrast, on the lower-noise dataset \agnews (19.0\% error rate), \tool attains its strong and stable performance over a broader range of $\alpha$ values, i.e., $\alpha \in [0.5, 0.8]$. This indicates that when the noise level is lower, \tool can effectively leverage both model predictions and previously refined labels.
Overall, \textit{these results empirically suggest combining model predictions with prior refined labels, while assigning greater weight to model predictions, i.e., a high value of $\alpha$, tends to yield better performance.}
\subsubsection{Dataset Size Analysis}

\begin{figure}
\centering
\begin{subfigure}{\columnwidth}
\centering
\includegraphics[width=\columnwidth]{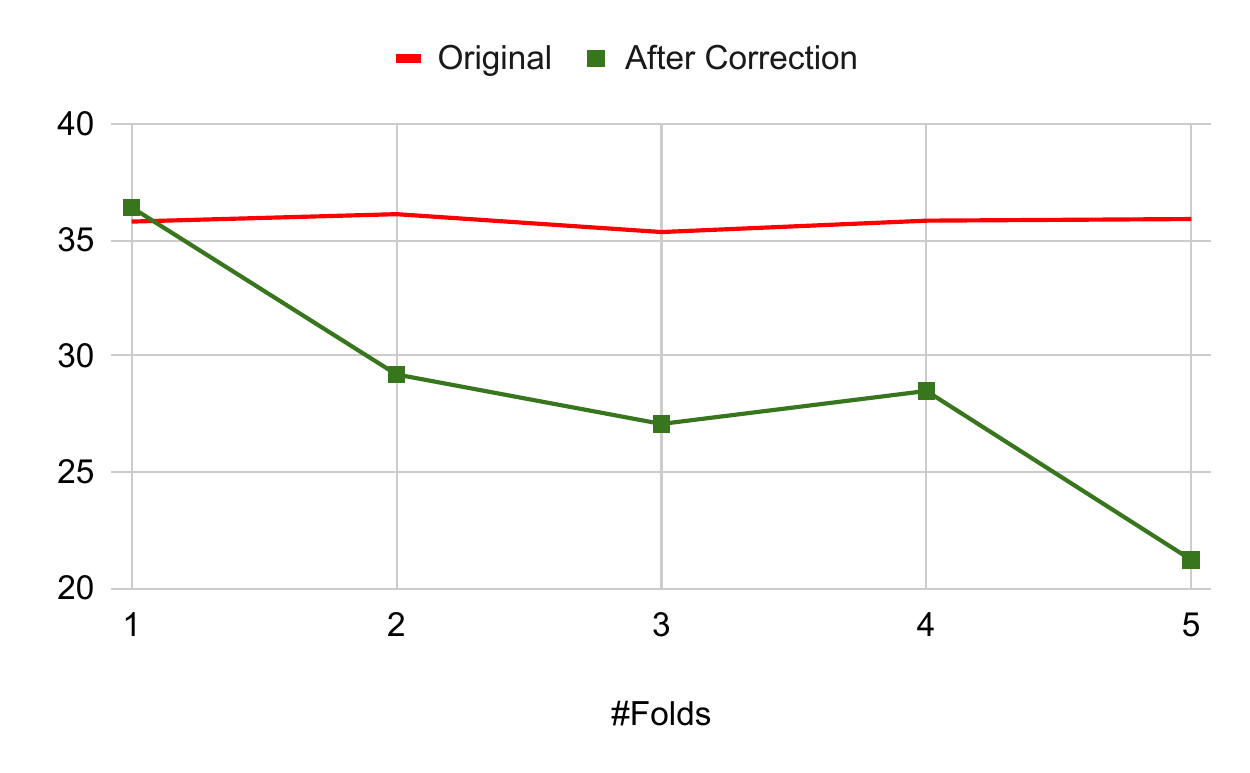}
\caption{\resisc}
\label{fig:resisc_sample_size}
\end{subfigure}\\

\begin{subfigure}{\columnwidth}
\centering
\includegraphics[width=\columnwidth]{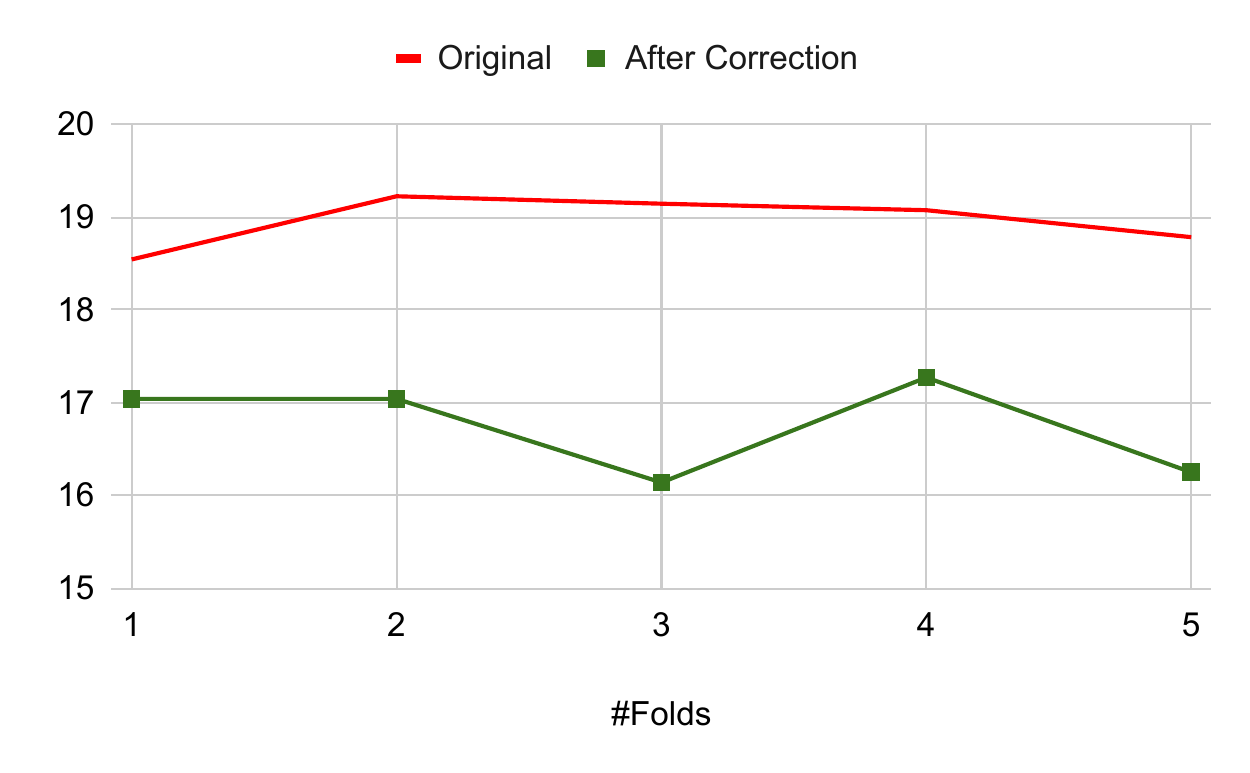}
\caption{\agnews}
\label{fig:agnews_sample_size}
\end{subfigure}

\caption{Impact of the sample size on \tool's performance, measured by \textit{Error Rate}}
\label{fig:sample_size}
\end{figure}

Figure~\ref{fig:sample_size} shows the impact of data size on \tool's performance. On both \resisc and \agnews, the error rate of the original datasets remains largely unchanged as the data size increases. This is expected, since mislabeled instances are approximately uniformly distributed across folds. Thus, adding more data introduces both clean and corrupted samples in similar proportions, leaving the overall noise rate relatively stable.

Overall, \textit{the effect of increasing data size on \tool's performance differs across datasets, reflecting their distinct noise characteristics.}
As shown in Figure~\ref{fig:resisc_sample_size}, on \resisc, increasing sample sizes significantly enhances the performance of \tool. When only two folds are used, \tool reduces the error rate of the original dataset by approximately 19\%. As the number of folds increases to five, this relative reduction improves substantially to 41\%.
However, the performance trend on \agnews (Figure~\ref{fig:agnews_sample_size}) exhibits fluctuations as the data size increases. For example, \tool achieves a larger noise reduction with three folds than with either two or four folds of data.

\begin{gtheorem}
\textbf{Answer to RQ4}: Although \tool is influenced by factors such as embedding models, the number of warm-up epochs, the number of correction iterations, and blending coefficient $\alpha$, it is robust to a wide range of parameter settings. The performance trends are stable and consistent across datasets, suggesting that \tool does not rely on fragile parameter tuning. This robustness indicates strong generalization ability under varying noise levels and dataset characteristics.
\end{gtheorem}

\subsection{Time Complexity Analysis}

In this work, all experiments were conducted on a server running Linux 6.6.105+ equipped with two NVIDIA T4 GPUs. 
As a common preprocessing step, all approaches require computing instance embeddings. On average, embedding a single instance takes approximately 0.02 seconds, and this cost is shared across all methods.

\tool demonstrates computational efficiency in detecting and correcting corrupted labels. It requires approximately 14 minutes to process a dataset. Specifically, \tool achieves the fastest runtime on the \yahoo dataset, completing label correction for 10K instances in about 8 minutes. The longest runtime occurs on \resisc, where processing 18.9K images takes around 30 minutes. This increase is mainly attributed to the larger dataset size.

Among the baselines, \semi is the most efficient, requiring only about 1 minute per dataset, due to its lightweight reliance on the $K$-NN algorithm. \selc exhibits moderate computational cost, taking about 21 minutes on average per dataset, with its longest runtime (around 30 minutes) observed on \organ.
\sidyp is the most time-consuming approach, requiring between 60 and 180 minutes per dataset, significantly slower than all other methods.
These results indicate that \tool achieves a favorable balance between computational efficiency and correction effectiveness. It remains practical for medium-scale real-world datasets while providing substantially stronger correction performance than faster but less expressive baselines.

\subsection{Threats to Validity}

The main threats to the validity of our work consist of internal, external, and construct validity threats.

\textbf{Threats to internal validity:} One potential threat lies in implementation correctness and hyperparameter selection. To mitigate this threat, we carefully reviewed our code. In addition, the hyperparameters are carefully selected via multiple experiments. 
We also make our code available publicly~\cite{website}, allowing other researchers to double-check and reproduce our experiments.

\textbf{Threats to external validity:} External threats concern the generalization of our findings to other datasets, domains, and noise conditions. To mitigate these threats, we evaluated our approach across diverse datasets and domains, under multiple noise conditions introduced by different automated labeling techniques. 
Moreover, we also evaluated \tool under a real-world noisy dataset (\clothing). 
In future work, we plan to evaluate \tool in additional domains, larger-scale datasets, and more complex labeling scenarios.

\textbf{Threats to construct validity:} A potential threat lies in the chosen metrics. To reduce this threat, we evaluated \tool and the baselines on two complementary metrics, 
\textit{error rate}, which directly reflects label correction accuracy, and \textit{downstream model performance}, which measures the practical utility of corrected datasets in supporting effective model training. Another threat is that automated labeling pipelines used to construct experimental datasets may not fully capture all forms of real-world noise. To address this issue, we employ diverse labeling techniques and additionally evaluate on a real-world noisy dataset (\clothing).
\section{Related Work}
\label{sec:related-work}

\textbf{Automated data annotation:}
Data annotation is a crucial step in developing high-quality ML/DL models~\cite{schmarje2022one}. To reduce the cost and effort of manual labeling, a variety of automated and semi-automated annotation approaches have been proposed. 
\textit{Weakly supervised learning} techniques~\cite{snorkel, hou2024kgred, galhotra2021adaptive} typically rely on heuristic rules, keyword matching, decision trees, or other expert-defined labeling functions to assign labels to data instances. In practice, multiple labeling functions are often applied, and their outputs are aggregated by a predefined or learned aggregation strategy to produce final labels. 
\textit{Semi-supervised learning} (SSL) further exploits both labeled and unlabeled data, aiming to infer or propagate labels from a small labeled subset to a larger unlabeled corpus. In these approaches, labels for unlabeled instances are automatically inferred using techniques such as label propagation, manifold regularization, or model-based prediction~\cite{label_spreading, bengio2006label, belkin2006manifold, lee2013pseudo, yang2022survey, song2024optimal, jiao2024learning}.
More recently, \textit{LLMs have been integrated into data labeling pipelines}, enabling new paradigms that leverage their strong contextual understanding and broad world knowledge~\cite{zhu2024apt, llm-in-the-loop, insightpilot}. However, despite their promise, several studies~\cite{llm-label-quality, chatgpt-label} have raised concerns about the quality, consistency, and reliability of labels generated by automated techniques. 

\tool is complementary to these labeling approaches, as it can be applied as a post-processing step to detect and correct noisy annotations, thereby improving the overall quality of the resulting labeled datasets.

\textbf{Corrupted label detection:}
To improve the quality of training data, various approaches~\cite{simifeat, retrieval-based, noiserank, yu2023delving} have been proposed for detecting corrupted labels. These methods generally aim to identify mislabeled instances without requiring access to a clean label set.
For instance, Liu \etal~\cite{retrieval-based} introduce a retrieval-based solution that detects noisy labels by identifying inconsistencies among similar instances in the feature space.
NoiseRank~\cite{noiserank} formulates noisy label detection as a ranking problem using Markov Random Fields (MRF), where instances are ranked according to their likelihood of being mislabeled and progressively refined through iterative training and noise filtering. COLA~\cite{cola} further enhances detection accuracy by capturing both local and global relationships among instances to assess label correctness.

Other studies~\cite{kim2024learning, sent, data-fault-localization} leverage partial supervision or auxiliary signals to facilitate label noise detection.
For example, SENT~\cite{sent} transfers noise distribution information to a trusted clean subset and trains a classifier using model-based features to distinguish clean and corrupted labels. 
Yin~\etal~\cite{data-fault-localization} propose DFauLo, which reframes corrupted label detection as the fault localization problem. DFauLo generates mutation-based DNN variants of a trained model and localizes mislabeled instances by analyzing prediction inconsistencies across these mutants.

While these techniques are effective in identifying potentially mislabeled data, they do not explicitly repair the detected errors. In contrast, \tool not only detects corrupted labels but also repairs them through a unified process. By jointly identifying and refining noisy labels, \tool improves overall data quality without discarding potentially informative instances, thereby preserving dataset size and maximizing the utility of available supervision.

\textbf{Corrupted label correction:}
Several studies~\cite{sidyp, selc, docta, zhang2025efficient} have investigated the problem of correcting corrupted labels.
Existing label correction approaches can be broadly categorized into two groups based on the source of their correction signals: neighbor-based correction and model-based correction.

\textit{Neighbor-based correction} approaches~\cite{sidyp, docta, kong2020knn, lee2024fastsimifeat} infer repaired labels by exploiting the label consistency among similar samples in the feature space. 
For instance, \semi~\cite{docta} identifies unreliable annotations based on neighborhood disagreement and corrects them using majority voting or ranking-based strategies. Similarly, \sidyp~\cite{sidyp} generates candidate labels from neighboring instances, and further refines them using a diffusion-based model. 

\textit{Model-based correction} approaches~\cite{selc, zhang2025efficient, zheng2020error, xu2025revisiting} primarily rely on a model's own predictions as correction signals. 
Zheng \etal~\cite{zheng2020error} theoretically demonstrate that the predictions of a noisy classifier can be a good indicator for detecting and correcting mislabeled instances.
Lu \etal~\cite{selc} and Zhang \etal~\cite{zhang2025efficient} propose approaches that use predictions from early training iterations to progressively refine mislabeled instances by blending the original annotations with model predictions.

Different from these existing approaches, \tool builds a noise-aware model and leverages its predictions to perform label correction in a more reliable and stable manner. Specifically, \tool estimates the underlying noise distribution and explicitly incorporates this information into the training process. 
Unlike prior model-based correction methods~\cite{selc, zhang2025efficient}, which rely on predictions from early training stages when model behavior can be unstable, \tool updates labels only after the model's training loss has stabilized. By explicitly modeling label noise and performing cautious, iterative soft relabeling, \tool provides a more principled and robust mechanism for repairing corrupted labels, particularly under high noise rates and complex noise patterns.

\textbf{Robustness to Label Noise:}
In addition to detecting and correcting corrupted labels, numerous approaches have focused on developing learning algorithms that are inherently robust to label noise~\cite{chenglearning, song2019selfie, li2019learning, noise-modeling,self-learning-with-noise, lienen2024mitigating}. These methods typically involve designing noise-tolerant model architectures or incorporating learning techniques that enable models to capture meaningful patterns from imperfect supervision while reducing the risk of overfitting to noisy labels. Unlike these \textit{model-centric} approaches, which focus on enhancing model robustness without modifying the data, \tool adopts a \textit{data-centric} perspective by explicitly improving the quality of the training data itself. In practice, \tool can be seamlessly integrated with model-centric methods to form a complementary pipeline that enhances both data reliability and model robustness.
\section{Conclusion}
\label{sec:conclusion}

This paper presents \tool, a novel framework for automatically correcting corrupted labels in noisy datasets. 
\tool explicitly models the underlying noise structure of the data and incorporates this information into the training of a noise-aware model, whose predictions are leveraged to progressively refine the incorrect labels. By estimating a noise transition matrix and performing iterative soft relabeling, \tool mitigates error propagation and enables stable repair of corrupted labels. 

Extensive experiments on six widely used image and text classification datasets demonstrate that \tool consistently outperforms SOTA label correction methods. On average, it reduces dataset error rates by approximately 25\%, with substantially larger gains under severe noise conditions. 
Furthermore, by producing cleaner training data, \tool leads to significant gains in downstream model performance.
Compared to noise-robust \textit{model-centric} approaches, \tool achieves superior downstream accuracy. These results highlight the practical advantage of data-centric strategies that explicitly clean the dataset over relying solely on robust training algorithms.

\printcredits
\bibliographystyle{elsarticle-num}
\bibliography{references}

\end{document}